\documentclass[conference]{IEEEtran}
\IEEEoverridecommandlockouts

\usepackage{cite}
\usepackage{amsmath,amssymb,amsfonts}
\usepackage{algorithmic}
\usepackage[dvipdfmx]{graphicx}
\usepackage{textcomp}
\usepackage{xcolor}
\usepackage{multirow}
\def\BibTeX{{\rm B\kern-.05em{\sc i\kern-.025em b}\kern-.08em
T\kern-.1667em\lower.7ex\hbo}x{E}\kern-.125emX}
     
\usepackage{flushend}
\usepackage{booktabs} 
\usepackage[ruled]{algorithm2e} 

\usepackage{soul}
\usepackage{url}

\providecommand{\keywords}[1]
{
  \small	
  \textbf{\textit{Keywords---}} #1
}

\begin{document}

\newcommand{\sys}{\textit{PassiFi}}
\newcommand{\sysAcc}{\textit{0.82}}

\title{Privacy-Preserving by Design: Indoor Positioning System Using Wi-Fi Passive TDOA}

\author{\IEEEauthorblockN{Mohamed Mohsen}
\IEEEauthorblockA{\textit{Shoubra Faculty of Engineering} \\
\textit{Benha University,
Benha, Egypt}\\
mohamed.hekal@feng.bu.edu.eg}

\and
\IEEEauthorblockN{Hamada Rizk}
\IEEEauthorblockA{
\textit{Osaka University
Osaka, Japan}\\
\textit{Tanta University
Tanta, Egypt}\\
hamada\_rizk@f-eng.tanta.edu.eg}
\and
\IEEEauthorblockN{Moustafa Youssef}
\IEEEauthorblockA{\textit{Dept. of Comp. Sc. \& Eng.,} \\
\textit{American University in Cairo, Egypt} \\
moustafa-youssef@aucegypt.edu}
}

\maketitle
\begin{abstract}

Indoor localization systems have become increasingly important in a wide range of applications, including industry, security, logistics, and emergency services. However, the growing demand for accurate localization has heightened concerns over privacy, as many localization systems rely on active signals that can be misused by an adversary to track users' movements or manipulate their measurements.
This paper presents PassiFi, a novel passive Wi-Fi time-based indoor localization system that effectively balances accuracy and privacy.
PassiFi uses a passive WiFi Time Difference of Arrival (TDoA) approach that ensures users' privacy  and safeguards the integrity of their measurement data while still achieving high accuracy. The system adopts a fingerprinting approach to address multi-path and non-line-of-sight problems and utilizes deep neural networks to learn the complex relationship between TDoA and location. Evaluation in real-world testbed demonstrates PassiFi's exceptional performance, surpassing traditional multilateration by 128\%, achieving sub-meter accuracy on par with state-of-the-art active measurement systems, all while preserving privacy.

\end{abstract}


\section{Introduction} \label{Introduction}
User's position has become one of the most valuable contexts in many applications like navigation, tracking, healthcare, emergency, etc \cite{buyukccorak2014indoor, emam2017adaptive, bahl2000radar}. The global positioning system (GPS) is considered the de facto standard technology for these applications and services in outdoor environments. However, when it comes to indoors where users to localize spend most of their time, GPS suffers from signal interference, reflections, and non-line of sight (NLoS) to the reference satellites\cite{aly2017accurate}. As a result,  scientific research and industry have devoted immense effort towards developing ubiquitous and accurate alternative technologies for indoor settings\cite{jung2011tdoa, wang2013position, lanzisera2006rf, bahillo2010hybrid}.
In this vein, various technologies have been opportunistically adopted for indoor positioning purposes, such as Wi-Fi, Bluetooth, ultra-wideband (UWB), cellular, etc\cite{zafari2019survey, al2011survey, mautz2009overview}. Each of these technologies has advantages and resolutions that fit specific applications. Among these technologies, Wi-Fi has been widely leveraged due to its ubiquitous coverage, and widespread support (the majority of commercial off-the-shelf devices support the IEEE 802.11 standard\cite{jiang2014communicating, mtibaa2015exploiting, hilsenbeck2014graph}).  

Many methods have been used for localization including multilateration, and fingerprinting, that are applied to angle-of-arrival-, RSSI-, and time-based techniques\cite{mendoza2019meta, al2014comparative}. 
However, the combination between time-based techniques and fingerprinting methods has recently been given the most attention due to their promising performance. \cite{hashem2020deepnar, hashem2020winar, rizk2022robust}.
In time-based techniques, the distance between a mobile device (e.g., smartphone) and an Access Point (AP) is estimated by multiplying the time a signal took to propagate from the AP to the mobile device by the propagation speed of that signal. Different approaches have been proposed to measure the signal's propagation time, such as time of arrival (ToA)\cite{golden2007sensor, chan2006time}, time difference of arrival (TDoA)\cite{ens2014unsynchronized}, and round trip time (RTT)\cite{hashem2020deepnar}. Recently, IEEE 802.11-2016 standard introduced Fine Time Measurement (FTM) protocol which gives mobile devices the ability to estimate the RTT between the mobile device and the APs. FTM protocol has increasing support from commercial APs and mobile devices~\cite{rttSupport}, making time-based techniques a promising solution for providing high-accurate and easily deployable indoor localization systems.   
However, the RTT estimation procedure faces several challenges in indoor environments, including multi-path propagation, latency, as well as non-line-of-sight (NLOS) transmission. Thus, signals travel indirect paths, resulting in longer distance estimation.

Fingerprinting\cite{bisio2013energy, youssef2005horus, firdaus2019review, wang2016csi, hashem2020winar} method is widely adopted by researchers due to its relatively good performance.
It consists of two stages: the offline stage and the online stage. During the offline phase, a fingerprint database is built by collecting different signatures (fingerprints) of Wi-Fi signals at different reference locations covering the area of interest. The collected fingerprints are used to build/train a model that can estimate the user's location using the received signals at run-time. Different types of models have been adopted, e.g., deterministic model in \cite{bahl2000radar}, probabilistic model in \cite{youssef2005horus}, or machine/deep learning models\cite{abbaswideep, hashem2020deepnar, rizk2022robust}. Although probabilistic techniques can counter the inherent wireless signal noise in a better way than deterministic techniques\cite{al2014comparative}, they usually assume that the signals from different access points (APs) are independent to avoid the curse of dimensionality problems. Practically, this hypothesis causes information loss. Therefore, deep learning techniques are widely adopted due to their ability to learn the joint distribution of underlying signals received from different APs, resulting in a high localization performance\cite{abbaswideep, hashem2020deepnar, rizk2022robust}. During the online phase, measurements are taken at some point in the area and fed to the model to estimate the position of the mobile device. 

Fingerprinting can be employed by using RSS fingerprints\cite{abbaswideep}, RTT fingerprints\cite{hashem2020deepnar}, or using both of them\cite{rizk2022robust}. RSS-based fingerprinting usually suffers from RSS fluctuations, multi-path fading, interference, and obstacles. Unlike RSS-based approaches, RTT is more robust in the indoor environment, as it is resilient to many challenges including multi-path fading, signal attenuation, transmission power variation, and radio interference\cite{feng2022analysis}. 
Although RTT estimates face many challenges in indoor environments, RTT fingerprinting approaches\cite{hashem2020deepnar, hashem2020winar,rizk2022robust, Farah2022MagttLoc} provide higher accuracy, and stability compared to RSS-based techniques.
However, most existing RTT-based approaches perform an active measurement procedure 
\cite{hashem2020deepnar, hashem2020winar,rizk2022robust, Farah2022MagttLoc}. 
In active processes, the mobile unit needs to send out, or exchange signals in order to get information about the environment.  Being an active process, RTT violates the user's privacy, as the user initiates the measurement session, and also responds with acknowledgment frames (ACK) to the AP. As a result, an adversary can leverage this to estimate the user's position, or corrupt the measurement session\cite{schepers2022privacy}. 


In this paper, we present \sys{}, a cutting-edge indoor localization system that leverages Wi-Fi technology for fine-grained and privacy-sensitive location estimation. The system uses a novel passive measurement approach based on the FTM protocol to accurately calculate the time-difference-of-arrival (TDoA) of Wi-Fi signals, which enables location estimation.
However, implementing a reliable and accurate indoor localization system is a challenging task, particularly when preserving user privacy is a crucial requirement. \sys{} overcomes these challenges by introducing several novel techniques to mitigate the limitations of active RTT measurement, reduce the impact of multipath propagation, and enhance the accuracy of the TDoA-based location estimation.
In particular, the passive measurement approach used by \sys{} ensures that user privacy is protected, as it does not transmit any data that could be used to identify the user or the device. This makes the system immune by design to several attacks that are commonly faced by active RTT-based systems, such as manipulating users' measurements or estimating their position.
Furthermore, the system constructs a fingerprint map of TDoA values at different discrete points within the area of interest, which is used to train a deep learning-based model that captures the complex relationship between TDoA fingerprints and user location. This allows \sys{} to achieve a fine-grained location accuracy that surpasses the existing systems.

Evaluation held in a real-world testbed showed the ability of the system to obtain a consistent median location error of \sysAcc{} m. This accuracy surpasses the accuracy of the traditional multilateration technique by $128\%$ and also presents an applaudable sub-meter accuracy compared to state-of-the-art systems in the considered testbed.

Our contributions are fourfold: 
(1) We propose a deep learning-based indoor localization system that ensures location privacy by employing a passive measurement approach that works on top of the IEEE 802.11mc standard and does not require any new hardware or protocol changes.
(2) We designed the system to combat the signal reflection, multi-path component effect, and non-line of sight challenges that face TDoA systems by fingerprinting  TDoA measurements at different reference points in the area of interest without advertising of user device. Then, these fingerprints are leveraged to train a deep model to capture the complex relation between TDoA fingerprints and the user location without overfitting.
(3) We studied the possible attacks that could negatively affect the performance of the active measurement localization systems and showed how our system is designed to invalidate them while keeping the localization performance. (4) we experimentally evaluate the performance of the proposed approach demonstrating its capability to localize without involving the user device in the measurement process and withstand variations in network and environmental settings.  

The rest of the paper is organized as follows: Section \ref{Related Work} discusses related work. Section \ref{Background} gives a brief explanation of the Wi-Fi FTM protocol. Section \ref{Active Procedure Privacy Violation} discusses the privacy threats associated with the active localization procedure. The basic idea behind \sys{} is discussed in Section \ref{The Basic Idea}.  In Section \ref{The System Details}, an overview of the system's building blocks is introduced, and the modules of the system are discussed in detail. We evaluate the system performance in Section \ref{Evaluation}. Finally, Section \ref{Conclusion} concludes the paper and presents future work.

\section{Related Work} \label{Related Work}
Research topics related to indoor localization have given considerable attention to Wi-Fi-based 
localization systems, as they use the same infrastructure of WLAN systems, and require no additional cost \cite{rizk2022robust, hashem2020winar, hashem2020deepnar, abbaswideep, eleryan2011aroma, abdellatif2013greenloc, neishaboori2013energy, mtibaa2015exploiting}. In the rest of this section, we will discuss RSS-, and time-based Wi-Fi indoor localization systems.

\subsection{RSS-based Systems}
The received signal strength (RSS) is an indicator of the power of the signal received by a wireless device. In propagation-based techniques \cite{mendoza2019meta, al2014comparative}, the RSS can be passively measured by the user equipment (UE) and used to estimate the distance between the UE and the access point (AP). Knowing the distance between the UE and AP and the positions of APs, the UE can be located anywhere on the circle whose center is an arbitrary AP, and its radius is the distance between the AP and the UE. To estimate the UE location, at least three RSS measurements from different APs to estimate the UE location as the intersection point of the three circles\footnote{This approach is referred to as trilateration or multilateration.}. 

However, RSS-based techniques suffer from inaccurate distance estimation and thus poor accuracy due to the non-line of sight (NLOS), multi-path interference, fading, etc \cite{feng2022analysis, wu2009practical, kumar2009distance}. 
Many indoor localization systems have been proposed to overcome these challenges by fingerprinting RSS measurements \cite{bisio2013energy, youssef2005horus, firdaus2019review}. RSS fingerprinting works in two phases: the offline phase and the online phase. In the offline phase, fingerprints (RSS measurements from several APs) are collected at predefined reference points to build a fingerprint database. These fingerprints are used to build a model that can be queried in the online phase to provide an estimate for the UE's position.
Over the years, different deep learning-based localization systems, e.g.,~\cite{wang2017csi,wang2017cifi,abbaswideep,rizk2019device,rizk2021device,monodcell,learntosee,floorRizk,cellRise,soloHamada,rizk2020omnicells} have shown good localization performance due to their ability to train efficient models that learn the complex patterns in the data and automatically extract discriminative features. Several deep learning architectures have been proposed in indoor positioning, including Restricted Boltzmann Machines in   
DeepFi~\cite{wang2017csi}, a deep convolutional neural network in~\cite{wang2017cifi}, and stacked denoising autoencoders in~\cite{abbaswideep}. 
Liphi system \cite{rizk2020gain,10.1145/3539659} was designed to enable zero-overhead fingerprinting-based localization through automatic labeling of WiFi scans using transportable LiDAR devices. These devices has been also used in~\cite{ohno2023privacy,Okochion-the-fly2022,9941552} to enable privacy-preserving human tracking.

The commonality between these approaches is the use of RSS as a measurement for the positioning of the UE.
However, this leads to unstable performance due to the presence of interference, fading,  reflection, and device sensitivity\cite{feng2022analysis}.

\textit{On the contrary hand, \sys{} relies on a time-based technique where the time difference of arrival of two received signals is measured and recorded as a fingerprint for each reference point in the area of interest. This ensures the system's resilience to interference, fading, and device sensitivity. }

\begin{figure}[!tbp]
 \centering
 \begin{minipage}[b]{0.94\linewidth}
\includegraphics[width=\linewidth,height=6cm,]{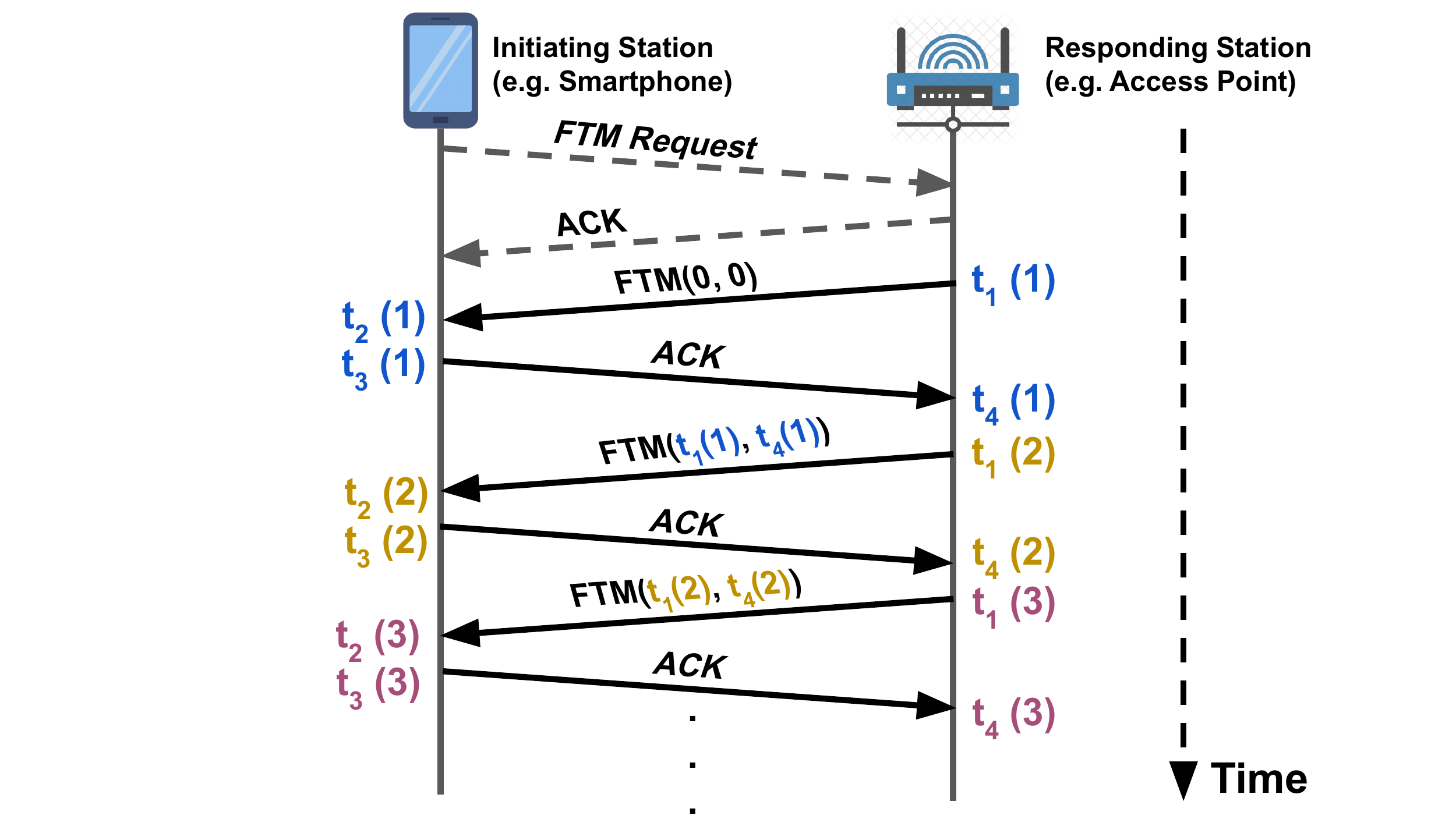}
    \caption{ The basic RTT estimation procedure in the IEEE 802.11mc standards.}
    \label{fig:FTM_proc}  
 \end{minipage}
\end{figure}

\subsection{Time-based Systems}

Time-based techniques leverage time information in order to estimate the UE's position. The time of arrival (ToA)\cite{golden2007sensor, chan2006time, lanzisera2006rf}, round trip time (RTT)\cite{hashem2020deepnar, hashem2020winar, ciurana2007ranging, gunther2005measuring, schauer2013potentials}, and time difference of arrival (TDoA)\cite{martin2020passive} are among the most popular techniques used for estimating user's position. 

Time of Arrival (ToA) is a one-way ranging technique where the time that the signal takes to arrive at the receiver is calculated by recording the arrival time of the signal at the receiver side and reading the transmit timestamp included in the transmit packet, then calculating the difference between them to get the time of flight (ToF). To accurately determine the signal's ToF,  strict synchronization between transmitter and receiver is mandatory in this process\cite{chan2006time, golden2007sensor, wang2013position}.

In the time difference of arrival (TDoA) technique, the receiving device calculates the difference in arrival time between two received signals from two different transmitters by timestamping the time of arrival of each received signal. Unlike ToA, TDoA does not require synchronization between transmitters and receivers, but strict synchronization is only essential among transmitters\cite{martin2020passive, zhang2006accurate, ens2014unsynchronized}.

On the other hand, Round Trip Time (RTT) is a two-way ranging process where the signal's time of flight (ToF) between transmitter and receiver is calculated by measuring the round trip time that a signal takes to travel from the receiver and return. Unlike ToA, RTT does not require synchronization between transmitter and receiver\cite{ciurana2007ranging, gunther2005measuring, rizk2022robust, hashem2020deepnar}.
Collecting enough number of these measurements in the area of interest enables estimating the user's location using Trilateration (or Multilateration). This method leverages the measured data in forming a finite set of equations that relates the position of the receiver and transmitters to the distance between the receiver and each transmitter where the position of the receiver is the only unknown\cite{schepers2022privacy}. The previous methods are negatively affected by any condition that causes an error in distance calculation like multi-path interference, non-line-of-sight, human body blockage, ... etc\cite{feng2022analysis, ibrahim2018verification}. To mitigate those challenges, time-based fingerprinting techniques have been adopted in \cite{hashem2020deepnar, hashem2020winar, rizk2022robust} to provide a more robust and highly accurate location determination process. 
{However, RTT measurement is considered an active procedure in which the user's device has to advertise itself. This makes the user more exposed to attacks by an adversary\cite{schepers2022privacy}. 

\textit{Unlike RTT, \sys{} preserves the privacy of the users without compromising on the system's performance. This is achieved as it provides the localization service without necessitating the user's device to announce its existence using the passive TDoA procedure.}
}

\begin{table}
\centering
\caption{Notations used in the paper.}
\label{table:notations}
{\renewcommand{\arraystretch}{1.1} 
\begin{tabular}{|p{1.2cm}|p{6.3cm}|l|}
\hline
\textbf{Notation} &\textbf{Description} \\ \hline 
$t_{1}(i)$ & The time of departure of the $i^{th}$ FTM frame recorded by the Responding Station during the burst. \\
$t_{2}(i)$ & The time of arrival of the $i^{th}$ FTM frame recorded by the Initiating Station during the burst.\\
$t_{3}(i)$ & The time of departure of the $i^{th}$ acknowledgment recorded by the Initiating Station during the burst.\\
$t_{4}(i)$ & The time of arrival of the $i^{th}$ The acknowledgment recorded by the Responding Station during the burst.\\
$t^{'}_{1}(i)$ & The time of arrival of the $i^{th}$ FTM frame (sent by Responding Station) recorded by Passive Station during the burst.\\
$t^{'}_{4}(i)$ & The time of arrival of the $i^{th}$ The acknowledgment (sent by Initiating Station) recorded by Passive Station during the burst.\\
$N$ & The burst size. \\
$c$  & The speed of Light ($3 \times 10^8$ m/s in space). \\
$\delta$ &Initiating Station processing time.\\
$RTT$ & The Round Trip Time.\\
$TDoA$ &Time Difference of Arrival.\\
$T_{ij}$ &Time of flight between station $i$ and station $j$.\\
$R_{ij}$ &Euclidean distance between station $i$ and station $j$.\\
$n$ & The number of the Responding Stations in the area.\\
$v$ & The number of captured TDoA samples at some location in the Online Phase.\\
$l_i$ & The  estimated 2-D coordinates $(x_i, y_i)$ of the $i^{th}$ sample in the $v$ collected samples.\\
$d_i$ & The inter-location distance which is  the distance between the $i^{th}$ location estimate and the rest of the estimated locations.\\
$u$ & The number of outlier-free location estimates as obtained from the Outlier Rejection stage in the Smoothing module during the Online phase.\\
$l^{*}$ & The center of mass of the outlier-free location estimates.\\
\hline 
\end{tabular}
}
\end{table}
\section{Background on Wi-Fi FTM} \label{Background}

The round-trip time is a time-based technique used to estimate the distance between two devices. It has been recently supported by the development of the FTM protocol in the IEEE 802.11mc standard.
The round-trip-time estimation procedure supported by IEEE 802.11mc involves two stations: the initiating station and the responding station. The former is the mobile unit to be positioned (e.g., smartphone), whereas the latter is the device (e.g., AP) to which the calculated distance is referenced. 
The whole RTT estimation procedure is illustrated in Figure~\ref{fig:FTM_proc}. First, the initiating station launches a measurement session by sending an FTM request frame to the responding station. The main purpose of this is to set up some parameters, such as the number of bursts, the number of measurements per burst, etc, {the burst size is the number of FTM frames sent by the responding station to the initiating station in the measurement procedure,} then the responding station replies with an acknowledge frame (ACK), and the session is activated. The responding station starts sending the first FTM frame and records the departure time of the frame (i.e., $t_{1}$ Figure~\ref{fig:FTM_proc}). That frame is then received by the initiating station which records the frame's reception time (i.e., $t_{2}$ as illustrated in Figure~\ref{fig:FTM_proc}). To confirm the reception of the FTM frame, the initiating station replies with an ACK frame and records its time of departure (i.e., $t_{3}$ as shown in Figure~\ref{fig:FTM_proc}). Once the ACK frame is received by the responding station, its arrival time is recorded (i.e., $t_{4}$). Finally, the timestamps $t_{1}$ and $t_{4}$ are included in the next FTM frame sent by the responding station to the initiating station in the same burst. For each single measurement in the burst, the initiating station is able to calculate several RTT relative to the responding station as follows:

\begin{equation} \label{RTT_calc}
    \mathrm{RTT}=\left(t_4-t_1\right)-\left(t_4-t_1\right)
\end{equation} But during the burst, several RTT values are collected relative to the same responding station then averaged out as follows:

\begin{equation} \label{RTT_avg_calc}
    AvgRTT= \frac{1}{N}\sum_{i=1}^{N}( \left(t_4(i)-t_1(i)\right) - \left(t_3(i)-t_2(i)\right) ) 
\end{equation} Then the distance can be estimated as follows:

\begin{equation} \label{RTT_distance}
    distance=0.5 \cdot AvgRTT \cdot \mathrm{c}
\end{equation} Where c is the propagation speed of light in space ($3 \times 10^8$ m/s ).

Making enough RTT estimates with different responding stations, several methods, including multilateration, fingerprinting, etc, could be used to estimate the location of the initiating station.

RTT estimating procedure is considered an active technique in which the users have to advertise themselves by initiating the process or responding with acknowledgments. This makes the user exposed to being eavesdropped on and attacked by strangers\cite{schepers2022privacy}. 



\begin{figure}[t]
 \centering
 \begin{minipage}[b]{0.94\linewidth}
    \includegraphics[width=\linewidth,height=5.5cm,scale=10]{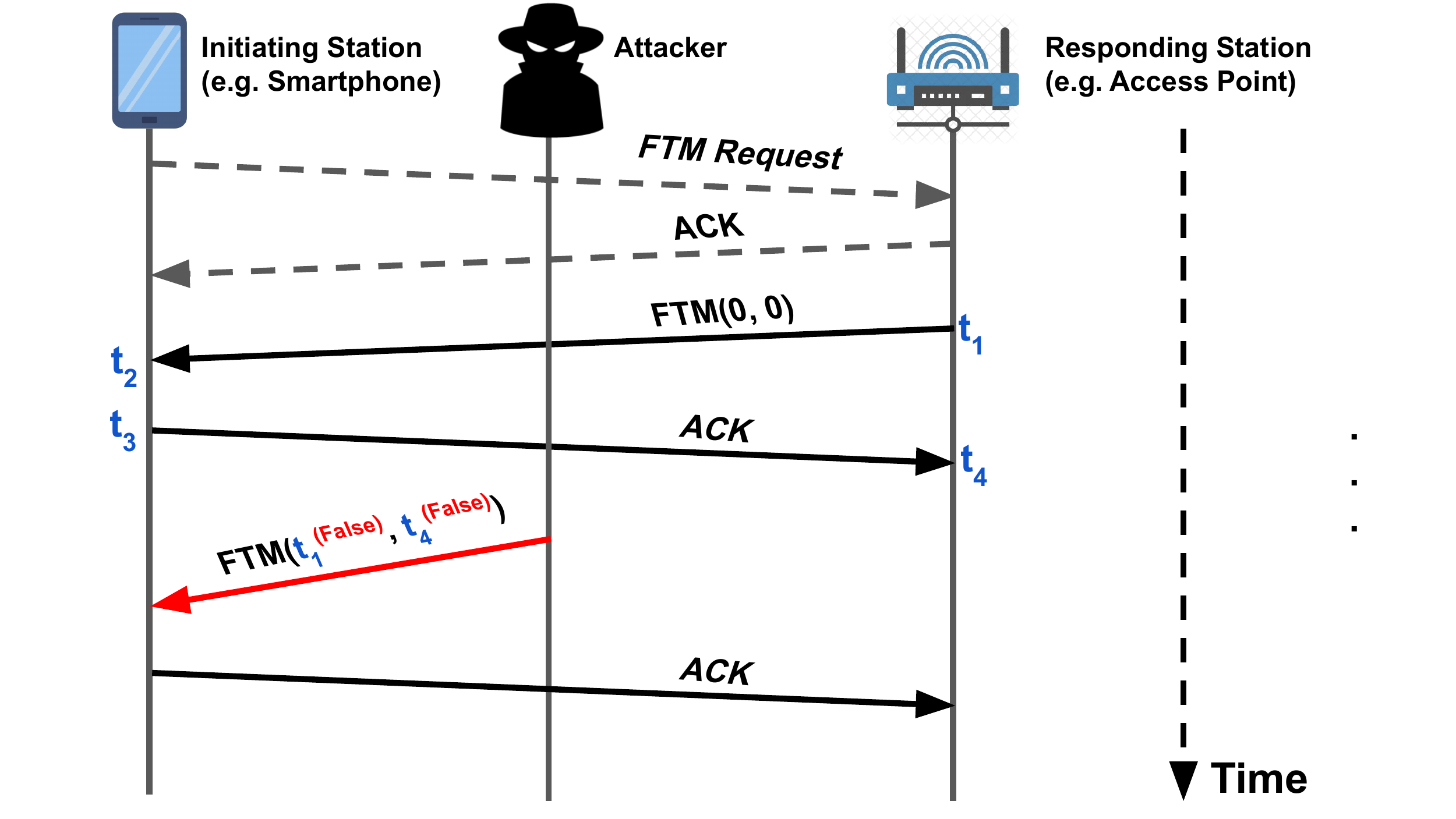}
    \caption{The attacker can manipulate the estimated distance by the user by spoofing FTM frames with falsified timestamps.}
    \label{fig:FTM_spoofing}  
 \end{minipage}
\end{figure}

\section{Threats to Location Privacy in Wi-Fi FTM} \label{Active Procedure Privacy Violation}
Wi-Fi FTM protocol is used to determine the distance between the user device and a hearable AP using round-trip time (RTT) (see eq.(\ref{RTT_avg_calc}) \& (\ref{RTT_distance})). Specifically, the user device initiates the process by sending an FTM request and then responds with ACK to each received FTM frame from the AP. 
In this vein, Wi-Fi FTM protocol is considered an \textit{active measurement procedure} in which the user device has to advertise itself, which gives  the adversary the chance to know their MAC address\footnote{Despite the availability of MAC address randomization techniques, they could be bypassed in several ways \cite{martin2020passive}, including power management side channel, correlation of sequence numbers, and fixed set of addresses.}, detect their existence then eavesdrop on it. 
The Round Trip Time (RTT) calculation, as outlined in equation (\ref{RTT_calc}), relies on two key components: the time difference ($t_4 - t_1$), recorded by the responding access point (AP) and transmitted unencrypted to the user's device, and the Initiator Processing Time (IPT) represented by ($t_3 - t_2$), recorded by the initiating device (such as the user's smartphone). If obtained by malicious actors, these values could be used to infringe on the user's privacy or manipulate their localization information, as discussed next.

\subsection{Threat to User Location Privacy} The adversary can use ($t_1$ and $t_4$), recorded and openly transmitted by the responding station (AP), to determine the user's location. Even though ($t_2$ and $t_3$), representing the Initiator Processing Time (IPT), are recorded by the user's device and not transmitted, an adversary can still estimate IPT by benchmarking the average IPT values of different Wi-Fi cards. This way, the adversary can determine the distance between the user's device and the AP and thus, learn the user's location \cite{schepers2022privacy}. This highlights the need for a localization system that can accurately estimate the user's location without compromising her privacy by advertising her presence.

\sys{} solves this problem by using Wi-Fi passive TDoA technique (see Fig.~\ref{fig:WiFi_TDOA}). The user device listens to the frames being exchanged between APs and records the time difference of arrival ($t_{4}^{'}$ - $t_{1}^{'}$) between these frames. The recorded information is then used to estimate the user's location using different algorithms. This technique is passive, as the user only listens to the frames, and the timestamps ($t_{4}^{'}$ and $t_{1}^{'}$) are dynamically changing based on the user's position. 
In this manner, attackers are unable to detect the user's existence or estimate their position, ensuring the privacy and security of the user's location information.

\subsection{FTM Frame Spoofing Attack} In this type of attack, the adversary manipulates the distance estimate received by the user device by tampering with the FTM frames \cite{schepers2021here}. The attacker begins by disrupting the original FTM frames sent from the responding station to the user device, followed by transmitting a spoofed FTM frame with fabricated timestamps ($t_{4}^{(false)}$ and $t_{1}^{(false)}$). The false timestamps are then used in the calculation of RTT, AvgRTT, and distance (eq. (\ref{RTT_calc}), (\ref{RTT_avg_calc}), and (\ref{RTT_distance})), leading to incorrect distance estimates and ultimately, a wrong estimation of the user's location (see Fig.~\ref{fig:FTM_spoofing}). This attack poses a threat to both the location privacy and the accuracy of the FTM measurements for users.





\subsection{ACK Frame Spoofing Attack}

In this attack, the adversary takes advantage of the known characteristics of ACK frames to manipulate the distance estimation of the user device \cite{schepers2021here}. The attacker sends a spoofed (fake) ACK frame to the responding station with a higher transmission power. This causes the responding station to record a falsified value of $t_4$, which represents the time of arrival of the ACK frame. This occurs because the devices rely on the highest power peak for Time of Arrival (ToA) estimation in the presence of multipath interference ( the same signal is received with different power levels and phases). As a result, the distance estimation becomes manipulated.

In contrast, \sys{} utilizes the fingerprinting method to passively collect TDoA data. The data collected only depends on the difference ($t_{4}^{'}$ - $t_{1}^{'}$) which is recorded by the passive station (i.e., the user device) and dynamically changes based on the user's movement. This enhances both the privacy of the users and the integrity of their measurements, making it resistant to FTM and ACK frames spoofing attacks.

\begin{figure}[t]
 \centering
 \begin{minipage}[b]{0.94\linewidth}
    \includegraphics[width=\linewidth,height=6.0cm,scale=10]{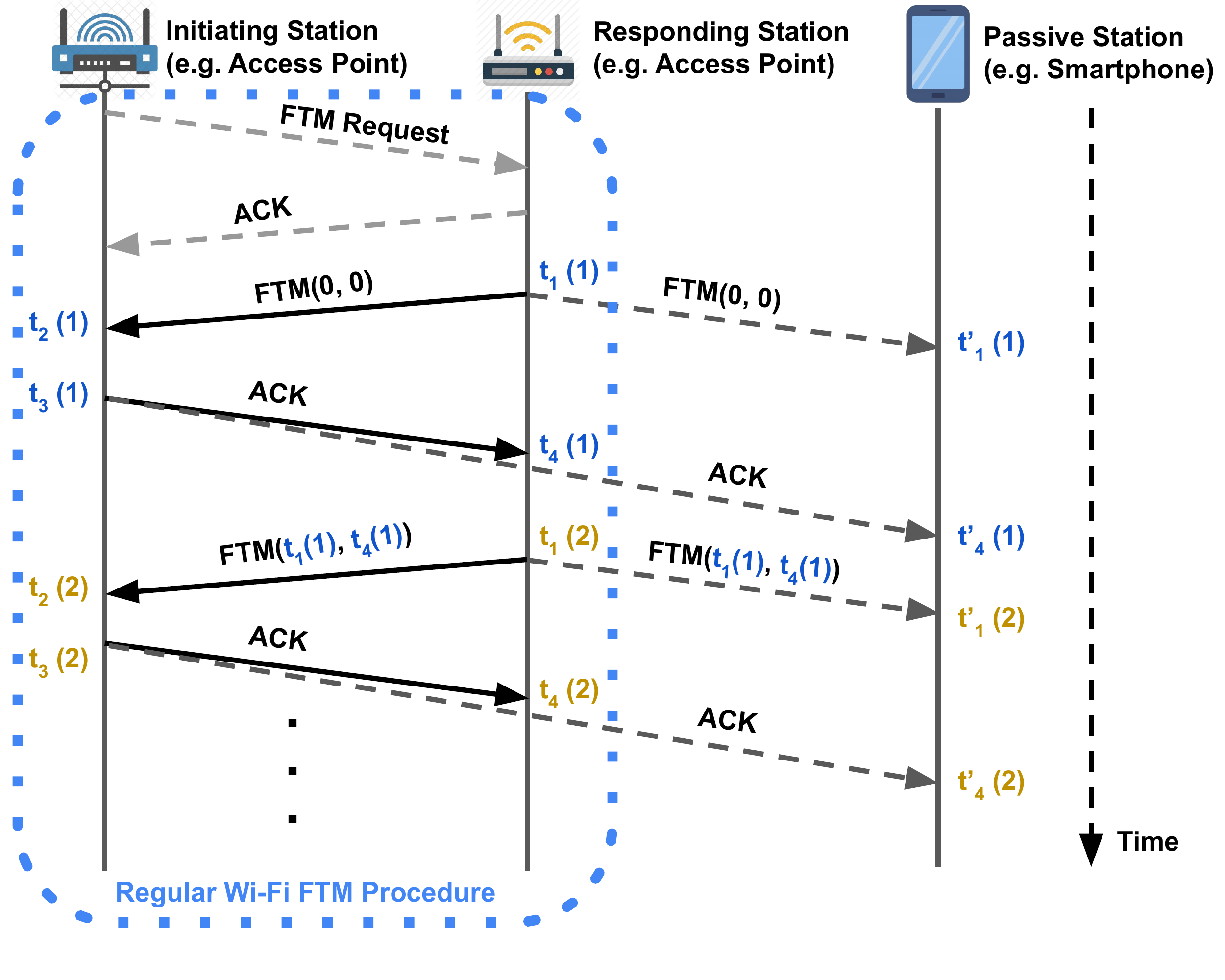}
    \caption{The estimation procedure of Wi-Fi Passive TDoA.}
    \label{fig:WiFi_TDOA}  
 \end{minipage}
\end{figure}

\section{The Basic Idea Behind \sys{}} \label{The Basic Idea}

\sys{} introduces a novel approach to inferring a user's position by leveraging the Wi-Fi FTM protocol. This method ensures that the user's presence is not advertised and the FTM protocol remains unchanged. The technique utilizes a passive TDoA approach, which is a spatial feature that varies with the user's position.

TDoA-based positioning techniques, as described in \cite{gustafsson2003positioning}, determine a user's location by computing the difference in time of arrival between signals received from two different access points. Unlike the traditional Wi-Fi FTM protocol, which involves only two types of stations (initiating and responding), the Wi-Fi passive TDoA method involves three types of stations: passive, initiating, and responding (as shown in Figure ~\ref{fig:WiFi_TDOA}).
The passive station, which is typically a device carried by the user, such as a smartphone, receives and records the time of arrival (ToA) of the received frames. The initiating station initiates the Wi-Fi FTM procedure with multiple responding stations. This allows the passive station to gather the necessary information to calculate the TDoA and determine its position without participating in the measurement, as described in \cite{martin2020passive}.
By using this passive TDoA method, \sys{} provides a secure and efficient solution for user positioning without compromising privacy or altering the standard Wi-Fi protocols.


The calculation procedure for TDoA is depicted in Figure ~\ref{fig:WiFi_TDOA}. The process begins with an FTM procedure initiated between the initiating station and the responding station. In this procedure (as described in Section \ref{Background}), the responding station sends an FTM frame to the initiating station. This frame is received by the passive station and the arrival time ($t_{1}^{'}$) is recorded. Subsequently, the initiating station sends an acknowledgment frame (ACK) to the responding station. This frame is also received by the passive station and its arrival time ($t_{4}^{'}$) is recorded. Finally, the TDoA can be calculated as follows:

\begin{equation} \label{WiFi_TDOA_eq_1}
TDoA = t_{4}^{'} - t_{1}^{'}
\end{equation}

In order to find the relation between the TDoA and the user's location, we follow two paths: 1) from the responding station to the passive station and 2) from the responding station to the initiating station, then to the passive station. Following these two paths, we get the following equation:
\begin{equation} \label{WiFi_TDOA_eq}
    TDoA + T_{r p}=T_{r i}+\delta+T_{i p}
\end{equation}
Where $T_{ab}$ is the time of flight of the signal from station a to station b, and $\delta = t_{3} - t_{2}$. 
Reordering eq.(\ref{WiFi_TDOA_eq}) to keep the information related to the passive station on the left-hand side:
\begin{equation} \label{WiFi_TDOA_eq_7}
    T_{i p} - T_{r p} = TDoA - T_{r i} - \delta  
\end{equation}
Adding and subtracting $T_{ri}$ from the right-hand side, we get:
\begin{equation} \label{WiFi_TDOA_eq_8}
    T_{i p} - T_{r p} = TDoA - 2 \cdot T_{r i} - \delta + T_{r i} 
\end{equation}
As the Wi-Fi FTM procedure is running between the initiating station and the responding station, then $RTT = (t_4 - t_1) - \delta = T_{ri} + T_{ir}$. But, $T_{ri}$ approximately equals $T_{ir}$, then:
\begin{equation} \label{WiFi_TDOA_eq_10}
    RTT = (t_4 - t_1) - \delta = 2 \cdot T_{ri}
\end{equation}
Using this, we can conclude that $(t_4 - t_1) = 2 \cdot T_{ri} + \delta$.
Substituting in eq.(\ref{WiFi_TDOA_eq_8}), we get:
\begin{equation} \label{WiFi_TDOA_eq_9}
    T_{i p} - T_{r p} = TDoA - (t_4 - t_1) + T_{r i} 
\end{equation}
Multiplying this equation by the speed of light $c$, we get:
\begin{equation} \label{WiFi_TDOA_eq_6}
    R_{ip} - R_{rp} = \Lambda_{RTT} + R_{ri} 
\end{equation}
where $R_{ab} = c \cdot T_{ab}$ is the Euclidean distance between station "a" and station "b", and $\Lambda_{RTT} = c \cdot (TDoA - (t_4 - t_1))$. 


\begin{figure}[t]
 \centering
 \begin{minipage}[b]{0.94\linewidth}
\includegraphics[width=\linewidth,height=4.5cm,scale=10]{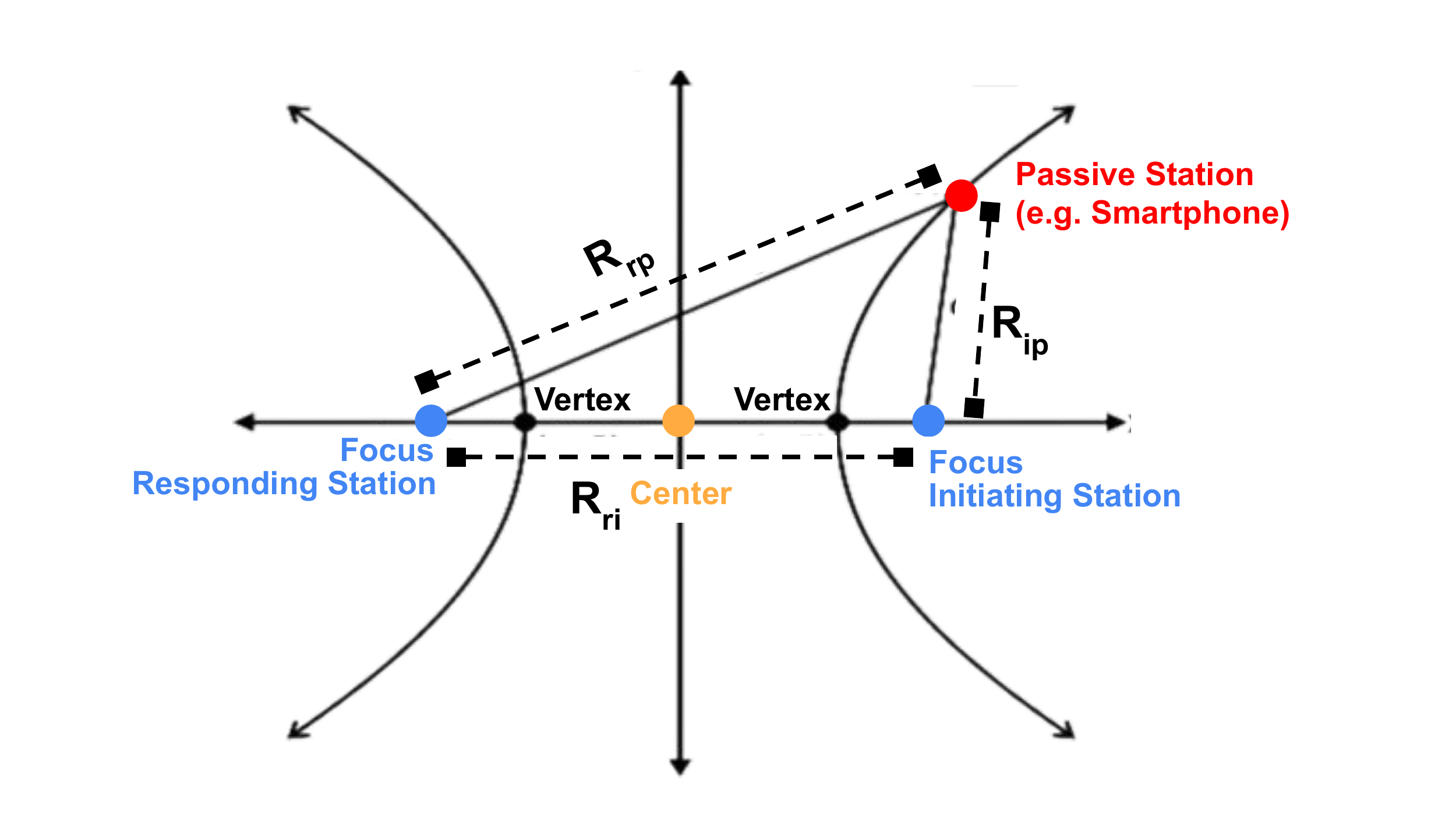}
    \caption{The locus of the passive station is hyperbola whose foci are the initiating and the responding stations.}
    \label{fig:tdoa_hyperbola}  
 \end{minipage}
\end{figure}

Eq.(\ref{WiFi_TDOA_eq_6}) represents the passive station location as a point on the hyperbola whose foci are the initiating and responding stations (same representation as in Fig.~\ref{fig:tdoa_hyperbola}). Collecting enough TDoA values relative to different responding stations, while keeping the same initiating station, the passive station location could be determined using algorithms such as multilateration. 

It is worth mentioning that in the Wi-Fi passive TDoA-based design, the users never advertise themselves, so it is impossible for an adversary to know their MAC address and detect their existence, so although $t_1$ and $t_4$ are sent in the clear, it is impossible for the attacker to estimate the values of $t^{'}_{1}$ and $t^{'}_{4}$ (see Fig. \ref{fig:WiFi_TDOA}).

Multilateration method is negatively affected by non-line-of-sight (NLoS) conditions, multipath interference, etc. So, the captured TDoA measurements are fingerprinted for training the deep learning model of \sys{}.



\section{\sys{} System Details} \label{The System Details}

\begin{figure}[t]
 \centering
 \begin{minipage}[b]{1\linewidth}
    \includegraphics[width=\linewidth]{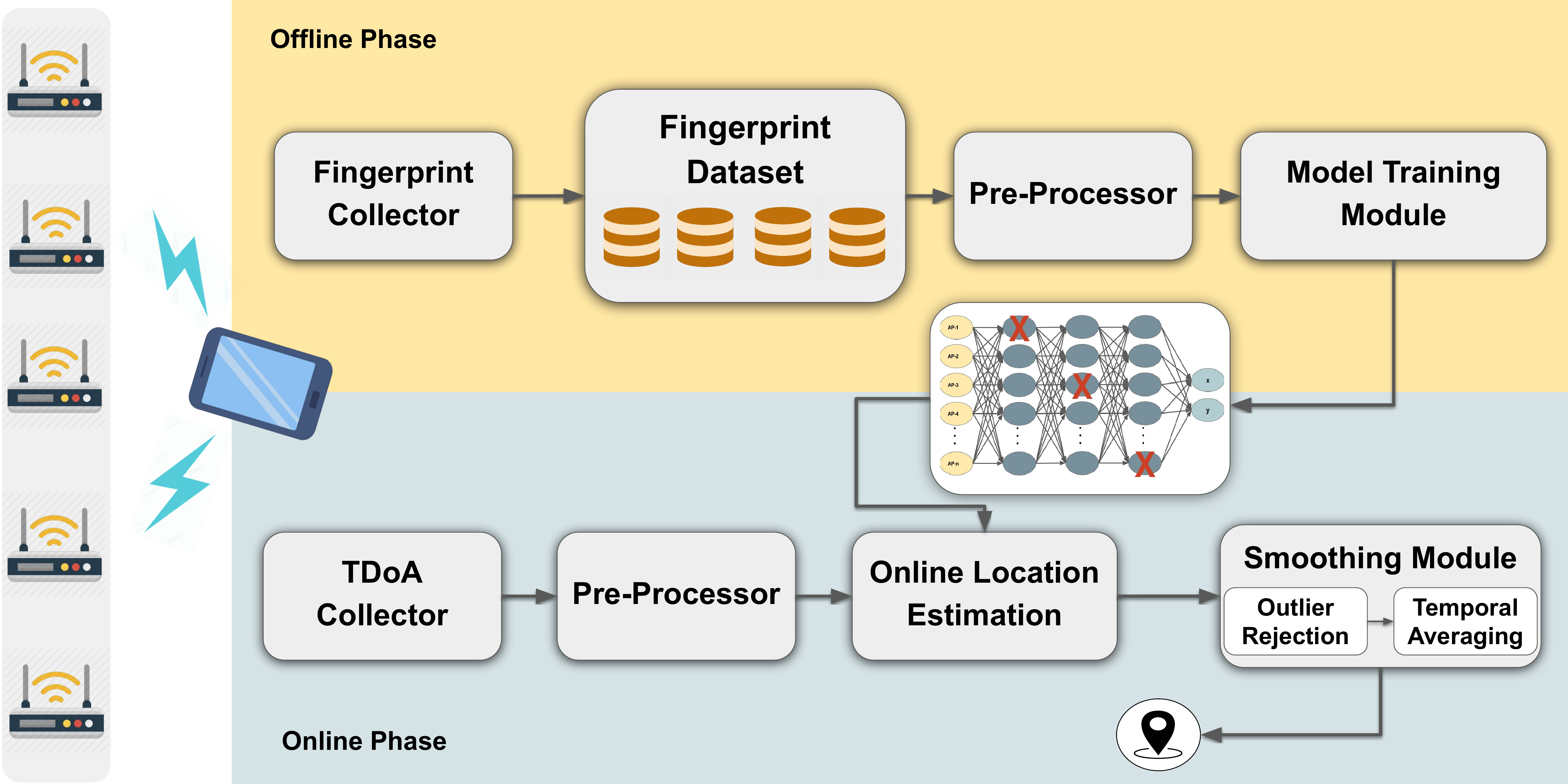}
    \caption{ \sys{} system Architecture.}
    \label{fig:Sys_Arch}  
 \end{minipage}
\end{figure}

\subsection{\sys{} System Overview}
The system architecture is shown in Figure~\ref{fig:Sys_Arch}. It works in two main stages: The offline training stage and the online localization stage.
During the Offline Training Stage, the area of interest is uniformly divided into discrete reference points where TDoA fingerprint data is collected using the \textbf{Fingerprint Collector} module (e.g., Android application) and it is done for once when the system is deployed for the first time. The \textbf{Pre-processing} module arranges the collected TDoA data overheard from APs by constructing fixed-size feature vectors, detecting the outliers, and performing normalization. After that, the pre-processed data is fed as input to the \textbf{Model Training} module. This module is responsible for training a deep neural network (DNN) model which is stored and used later in the online localization stage.
During the Online Localization Stage, the user location is estimated by feeding the preprocessed TDoA values relative to different responding APs to the trained deep-learning model. First, the collector app scans for APs, and collect the associated TDoA values. Then, the scanned data is processed and fed to the trained localization model. Finally, the \textbf{Online Location Estimation} module is used for further refining the estimated location obtained by the trained model.

\subsection{Pre-processing Module}
This module works during both the offline and online phases. It performs two steps: extending feature vectors and feature normalization. 
Due to noise in the wireless channel, not all responding stations are heard in all scans. The \textbf{Pre-processing} module ensures a consistent input size to the deep learning model by fixing/extending the length of the scanned input vectors. In case of not hearing a responding station, TDoA is given a value of 200 nanoseconds, which is equivalent to more than or equal to 60 meters far from that responding station. This value is larger than any TDoA value for the APs in the scanning range.

Then, the TDoA values are normalized to be in the range between [0, 1], as it improves the convergence time of the model\cite{kang2013stacked}. 

\begin{figure}[t]
 \centering
 \begin{minipage}[b]{0.94\linewidth}
    \includegraphics[width=\linewidth,height=6cm,]{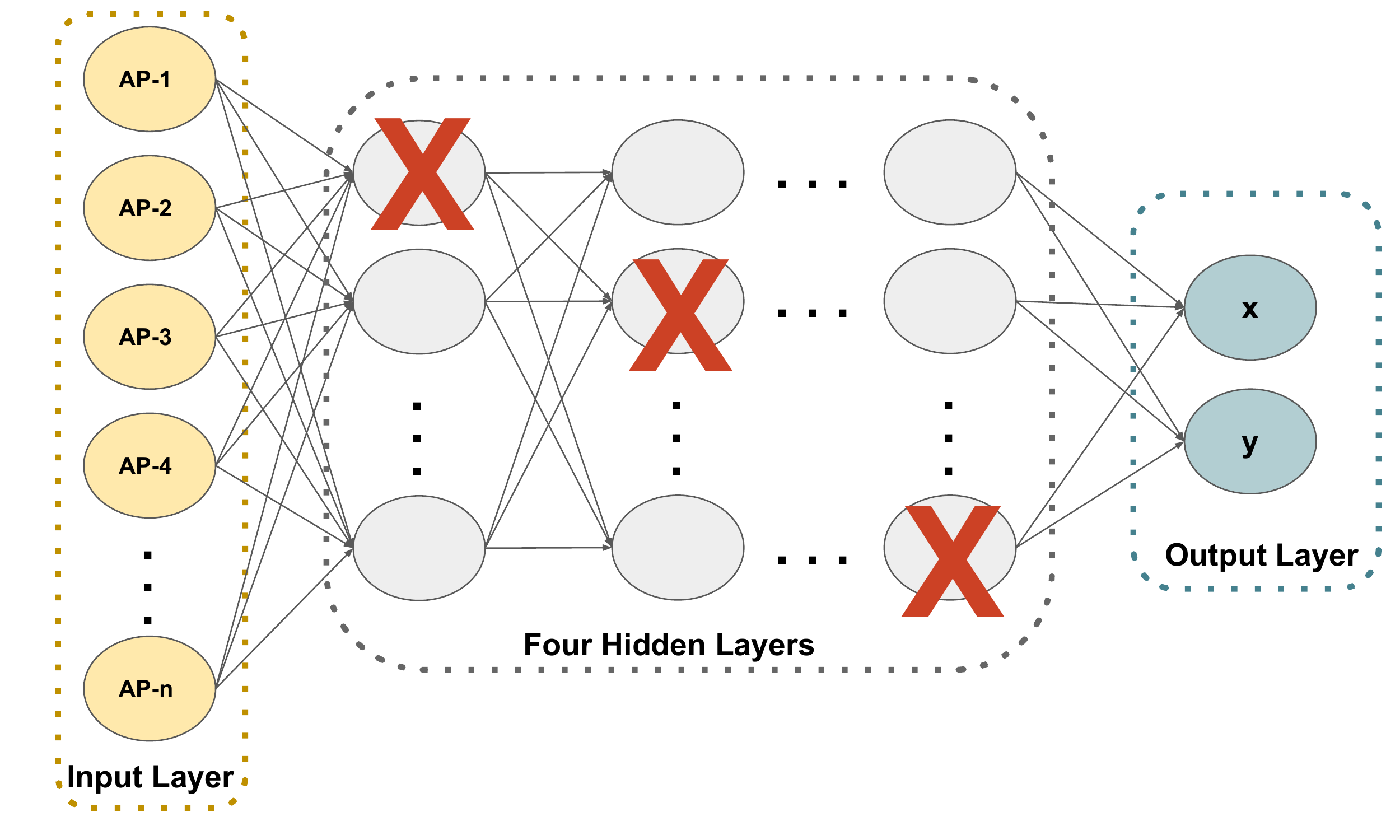}
    \caption{The model adopted by \sys{}. The input is the TDoA measurements of different APs (features). The output is the 2-D coordinates (x, y) of the location related to some TDoA vector.}
    \label{fig:DNN_model}  
 \end{minipage}
\end{figure}
\subsection{The Model Training Module}
This model is responsible for capturing the complex relationship between measured TDoA vectors and the location it is related to.
We used a deep-learning-based regression model. The model input is the pre-processed TDoA vectors collected at different reference points in the area of interest. The output is the 2-D coordinates (x, y) of the location corresponding to some TDoA vector.

Figure~\ref{fig:DNN_model} shows our DNN model architecture. The model consists of an input layer with the same size as the number of responding stations in the area, hidden layers, and a linear activated two-neuron output layer. Each hidden layer consists of 300 neurons that are fully connected with the previous layer and the next layer.

During the offline phase, the \textbf{Model Training} module is used to train the deep neural network model using the preprocessed TDoA fingerprint-scanned data at each reference point in the area of interest. The output is represented as a pair (x, y) which represents the 2-D coordinates of the reference points corresponding TDoA input vector.

To increase the model robustness and avoid overfitting during training, \sys{} employs two regularization techniques:
First, we used \textbf{dropout regularization} ~\cite{srivastava2014dropout} which has been shown to be useful for training deep networks. This is achieved by randomly removing (i.e., dropping out) some neurons from each layer in the network as well as their connections (Fig.~\ref{fig:DNN_model}).
Dropout has the effect of making the training process noisy by forcing neurons within every layer to randomly depend more or less on the inputs.  Hence, it prevents the neurons from relying on each other during training which results in making the model more robust to new data and less likely to memorize (i.e., overfit) the training data. We provide an experiment to show the effect of dropout on the results in Section~\ref{Evaluation}.
Second, \sys{} leverages early stopping to end the training process once improvements are no longer obtained after a definite number of epochs (patience value).

\subsection{The Online Location Estimation Module}
This module is used to estimate the unknown position of the user during the online phase, given the TDoA vector calculated at that position. We can estimate the user location by feeding the TDoA vector to the pre-trained model to get an estimate of the user location (x, y).

To further improve the system's stability to the abrupt changes in the measurements,
consecutive estimates are further processed to provide a fine-grained  estimate of the user location. Towards this end, \sys{} 
 employs two methods: \textit{Outlier Rejection} and \textit{Temporal Averaging}. Assuming ${v}$  TDoA vectors are fed sequentially to the trained model, and thus  ${v}$ estimates of the user location will be obtained. Ideally, the variance of those estimates should be near zero, which is not typically the case in practice due to the presence of multipath and different types of anomalies in the wireless channel. Thus, \sys{} lessens that variance and ensures the system performance by removing outlier candidates and then averaging the remaining ones, as described next.

\subsubsection{Outlier Rejection}
This technique removes the odd location estimates that may occur due to noisy wireless channel. To achieve that, \sys{} calculates the distance between each location estimate and the rest of the $v$ estimated locations (those distances are  called inter-location distances) as:
    \begin{equation} \label{inter_location_distance}
         d_{i} =\sum_{j \neq i} \frac{d\left(l_i, l_j\right)}{v-1}
    \end{equation}
Where $l_i$ is the coordinates ($x_i$, $y_i$) of $i^{th}$ point, and d($l_i$, $l_j$) is the Euclidean distance between the location $l_i$ and $l_j$. After that, \sys{} calculates the average inter-location distance $d_{avg}$
    \begin{equation} \label{inter_location_distance_avg}
         {d_{avg}}=\frac{1}{v}\sum_{i=1}^{v}d_i
    \end{equation}
and rejects all the location estimates whose inter-location distance satisfy $d_{i} > d_{avg}$\cite{rizk2015hybrid}.

\begin{figure}[t]
 \centering
 \begin{minipage}[b]{0.94\linewidth}
    \includegraphics[width=\linewidth,height=6cm,]{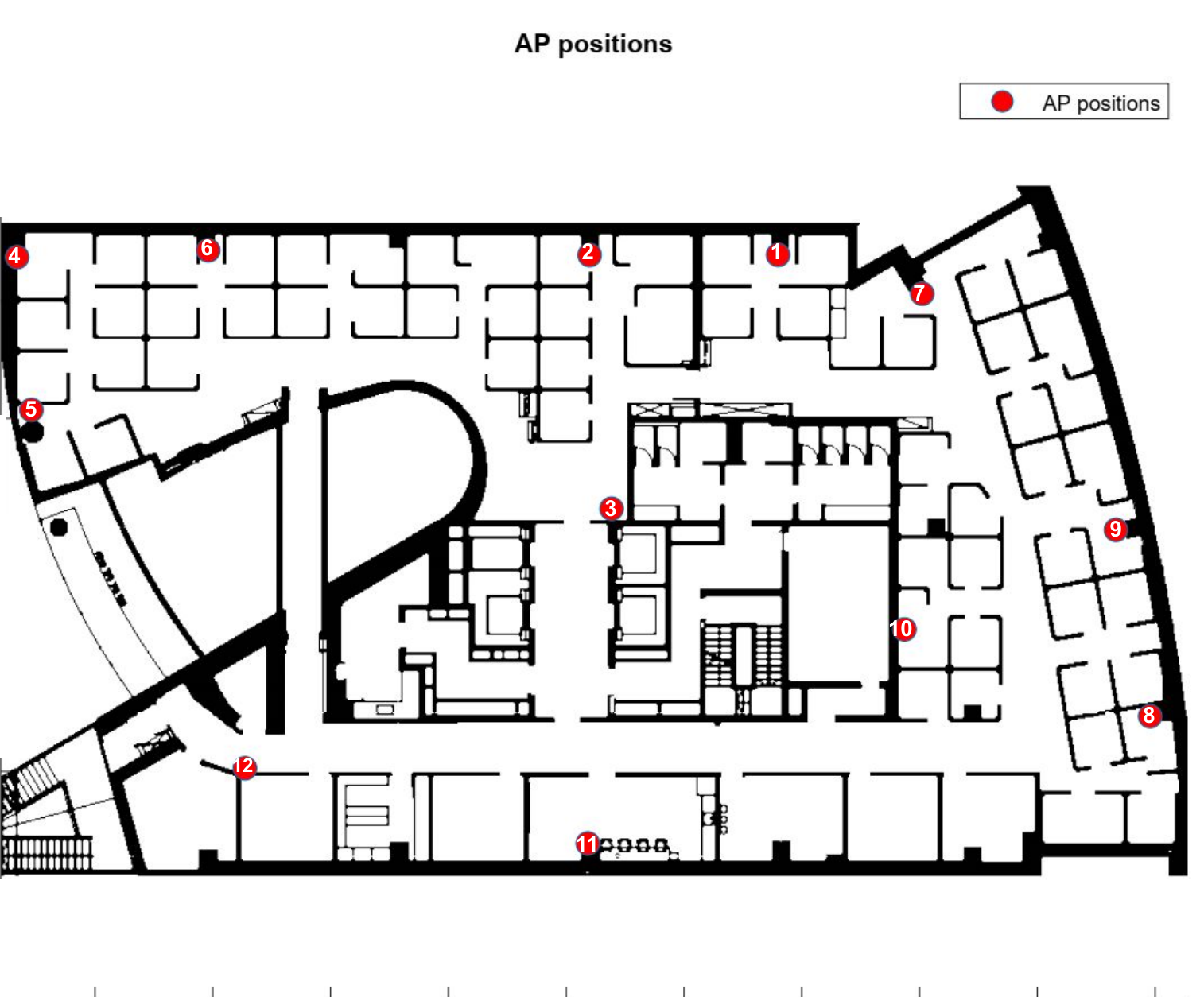}
    \caption{ Office Floor Plan with access points places in red.}
    \label{fig:Office_2}  
 \end{minipage}
\end{figure}

\begin{figure*}[t]
    \centering
    \begin{minipage}{0.32\linewidth}
        \centering
        \includegraphics[width=\linewidth, height=5.0cm]{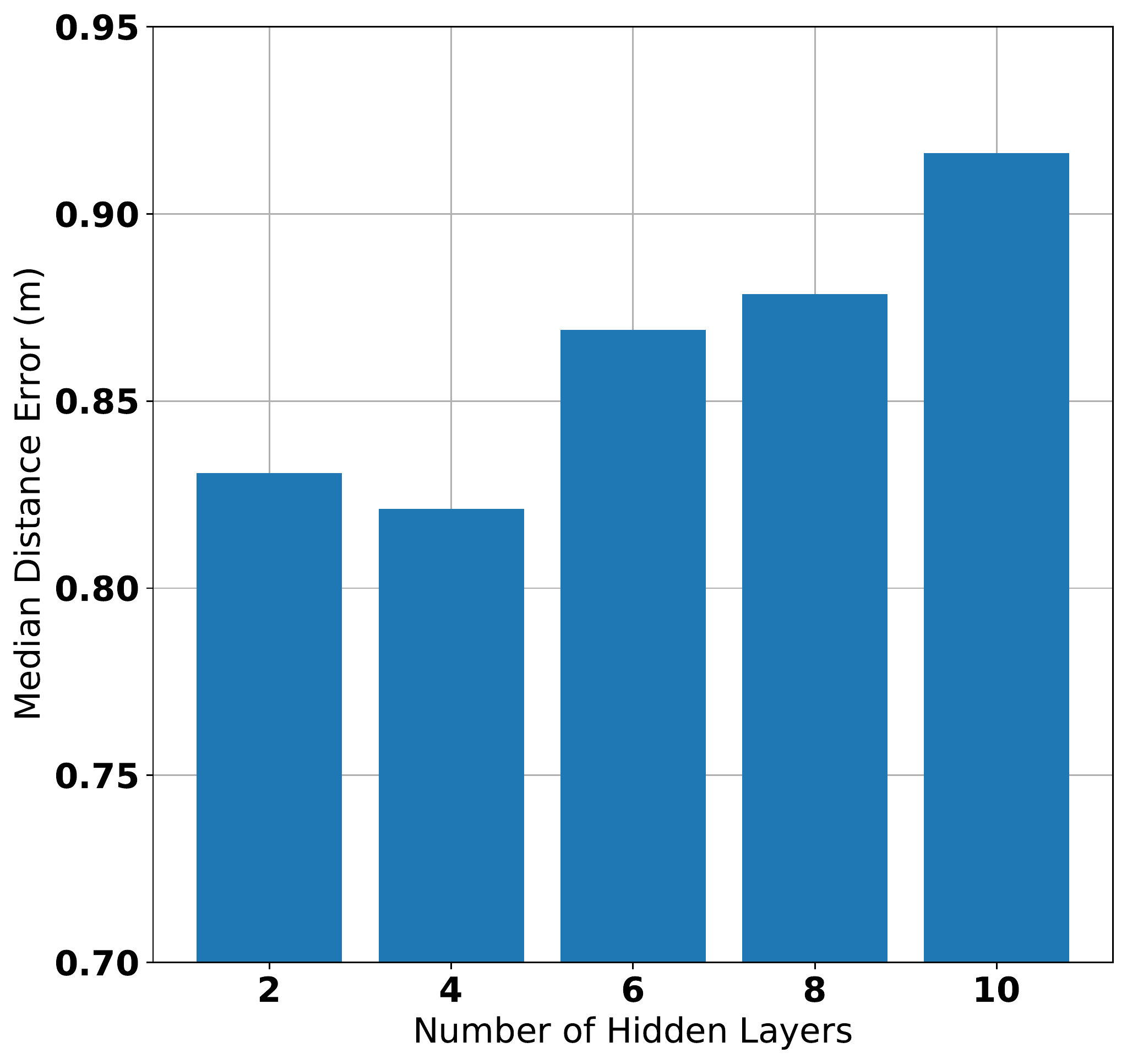}
        \caption{ Effect of changing the number hidden layers on system accuracy.}
        \label{fig:n_hidden_layers}  
    \end{minipage}
   \begin{minipage}{0.32\linewidth}
        \centering
        \includegraphics[width=\linewidth, height=5.0cm]{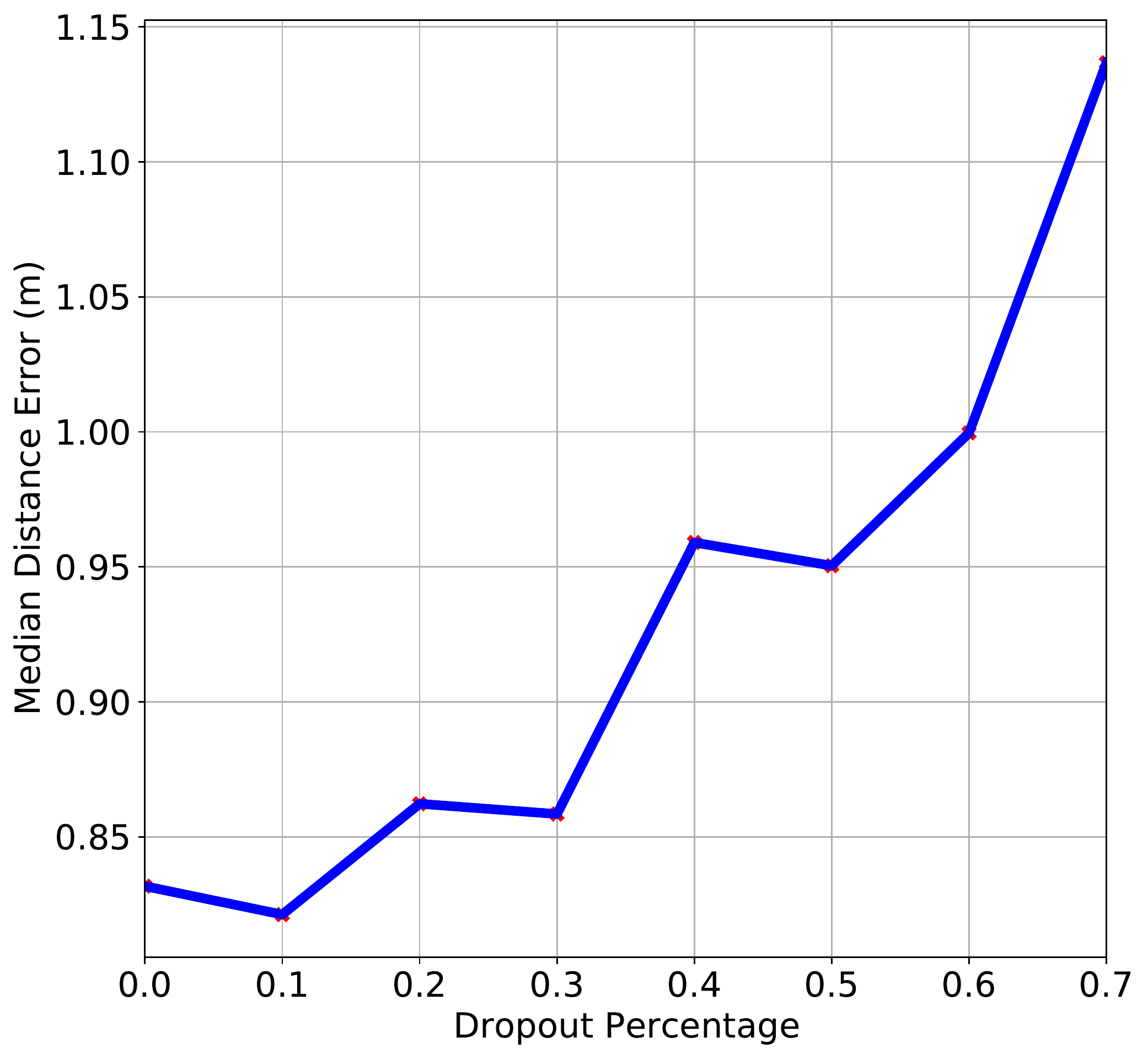}
        \caption{ Effect of the changing Dropout percentage on system accuracy.}
        \label{fig:dropout}  
    \end{minipage}
   \begin{minipage}{0.32\linewidth}
        \centering
        \includegraphics[width=\linewidth,height=5.0cm,]{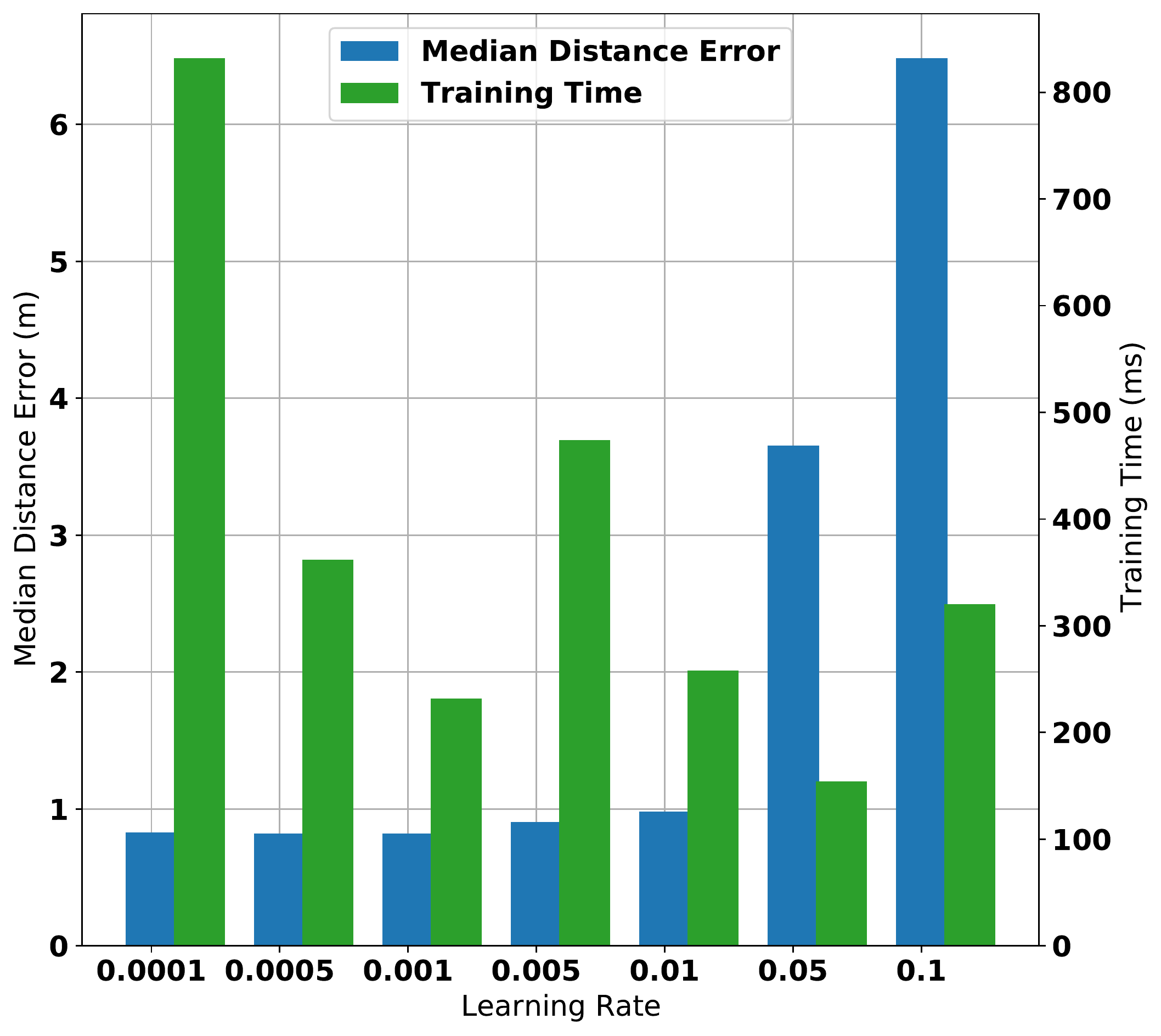}
        \caption{ Effect of changing the training learning rate on system accuracy.}
        \label{fig:learning_rate} 
    \end{minipage}
\end{figure*}
\subsubsection{Temporal Averaging}
This method further smooths the user location ($l^{*}$) by averaging the remaining outlier-free estimates. Assume we have $u$ samples, where $u \le v$, the average location could be calculated as follows: 
    \begin{equation} \label{location_avg}
         {l^{*}}=\frac{1}{u}\sum_{i=1}^{u}l_i = \frac{1}{u}(\sum_{i=1}^{u}x_i,\sum_{i=1}^{u}y_i)
    \end{equation}


\section{Evaluation} \label{Evaluation}
\subsection{Experimental Setup and Data Collection}

A large dataset \cite{h5c2-5439-20} is used to validate the performance of the proposed system. The data was collected at a real-world indoor testbed as summarized in Table~\ref{table:testbed}.
The layout of the testbed is shown in Figure~\ref{fig:Office_2}, which spans an area of $\sim1500m^2$. 
This area is an open space with a cubical office floor of a metal frame of  1.5m height, causing reflections of the transmitted signals. 
A structure of thick concrete walls at the center of the office. These walls form a constant NLoS propagation channel between the client and the APs in certain sections of the office. 
\begin{table}
        \centering
        \caption{Testbed Summary.}
        \label{table:testbed}
        \begin{tabular}{|c|c|c|}
        \hline
        \textbf{Parameter} & \textbf{Value}   \\ \hline
         
      \textit{ Area}& $\sim 50m \times 30m$  \\ \hline
      \textit{ Number of Responding Stations $(n)$}&  11  \\ \hline
      \textit{ Number of Initiating Stations}& 1  \\ \hline
      \textit{ Number of Training Points}& 3515  \\ \hline
      \textit{ Number of Testing Points}& 2343  \\ \hline

        \end{tabular}

\end{table}
    \begin{table}
            \centering
            \caption{Default parameters values used in evaluation.}
            \label{table:parameters}
            \begin{tabular}{|l|l|l|}
            \hline
            \textbf{Parameter} & \textbf{Range}  & \textbf{Default} \\ \hline
            \textit{ Dropout rate (\%)}& 0-70 & 10 \\ \hline
             
          \textit{ Number of Patience Epochs}& 1-50 & 50 \\ \hline
           \textit{ Number of Hidden Neurons }& 100-500 & 300 \\ \hline
           \textit{ Number of Hidden Layers }& 1-10 & 4 \\ \hline
           \textit{ Learning Rate($\alpha$) }& 0.0001-0.1 & 0.001 \\ \hline
           \textit{ Number of Reference Points}& 1-5858 & 5858 \\ \hline
           \textit{ Number of Responding Stations}& 1-11 & 11 \\ \hline
           \textit{ Number of Collected Samples per Point}& 10-100 & 100 \\ \hline
           \textit{ Number of Samples $(v)$}& 10-100 & 100 \\ \hline
            \end{tabular}

    \end{table}
The data was collected using a device with Intel® Dual Band Wireless- AC 8260 chip-set. The device was mounted on a robotic vehicle that collects data relative to different access points while moving over about 5858 points that cover the area of interest.

To effectively train the proposed model, our data augmentation technique~\cite{rizk2019effectiveness} is employed to the obtained TDoA vectors. This technique increases the amount of training data without extra data collection overhead by adding White Gaussian Noise (WGN) to TDoA values. \sys{} trains the deep model with the default hyperparameters summarized in Table \ref{table:parameters}. The system's accuracy represents the Euclidean distance (error) between the estimated location and the ground truth.



\subsection{Effect of Changing System Parameters}

\subsubsection{Number of Layers in the Deep Neural Network}
Figure~\ref{fig:n_hidden_layers} shows the effect of changing the number of hidden layers on \sys{} performance. As shown in the figure, the more hidden layers to consider, the better accuracy (i.e., less localization error) \sys{} can achieve until it reaches an optimal value at four layers. Beyond four hidden layers, the accuracy of the performance scores no further improvements. This can be justified due to two reasons. Firstly, increasing the number of layers increases the distributed learning ability of the localization model. In this vein, the model has enhanced computing power enabling the better fitting of the underlying function (without underfitting). Secondly, as few layers as four are enough to allow the localization model to learn the location of the mobile unit.  With very deep networks (adding more  layers), the model tends to overfit the training data reducing its flexibility and thus its accuracy. As a result, ~\textit{four} layers are set as the default number of layers in the \sys{} model to achieve a balance between underfitting and overfitting.

\subsubsection{Dropout Percentage}

 The effect of varying the percentage of dropout is shown in Figure \ref{fig:dropout}. It can be observed from the figure that at a dropout rate of 0.1, the best performance of \sys{} is achieved. This confirms the role of dropout regularization in boosting the generalizability of the trained model and ensures its resilience to over-fitting the training data. 
However, the model tends to under-fit the training data at larger dropout rates as many neurons are dropped off.

\subsubsection{Learning Rate$(\alpha)$}
Figure~\ref{fig:learning_rate} shows the effect of changing the learning rate on the performance of \sys{}. The learning rate controls how the model weights are updated while seeking a minimum loss during the optimization process. The figure shows that \sys{} obtains its best localization accuracy at a learning rate value of 0.001.
\sys{} localization accuracy is degraded at a larger learning rate as it may lead to a divergent training process (the gradients may overshoot). On the other hand, lower learning rates lead to approximately the same accuracy as the case where $\alpha=0.001$, but the training process converges  more slowly.

\begin{figure*}[t]
    \centering
    \begin{minipage}{0.32\linewidth}
        \centering
        \includegraphics[width=\linewidth,height=5.0cm,]{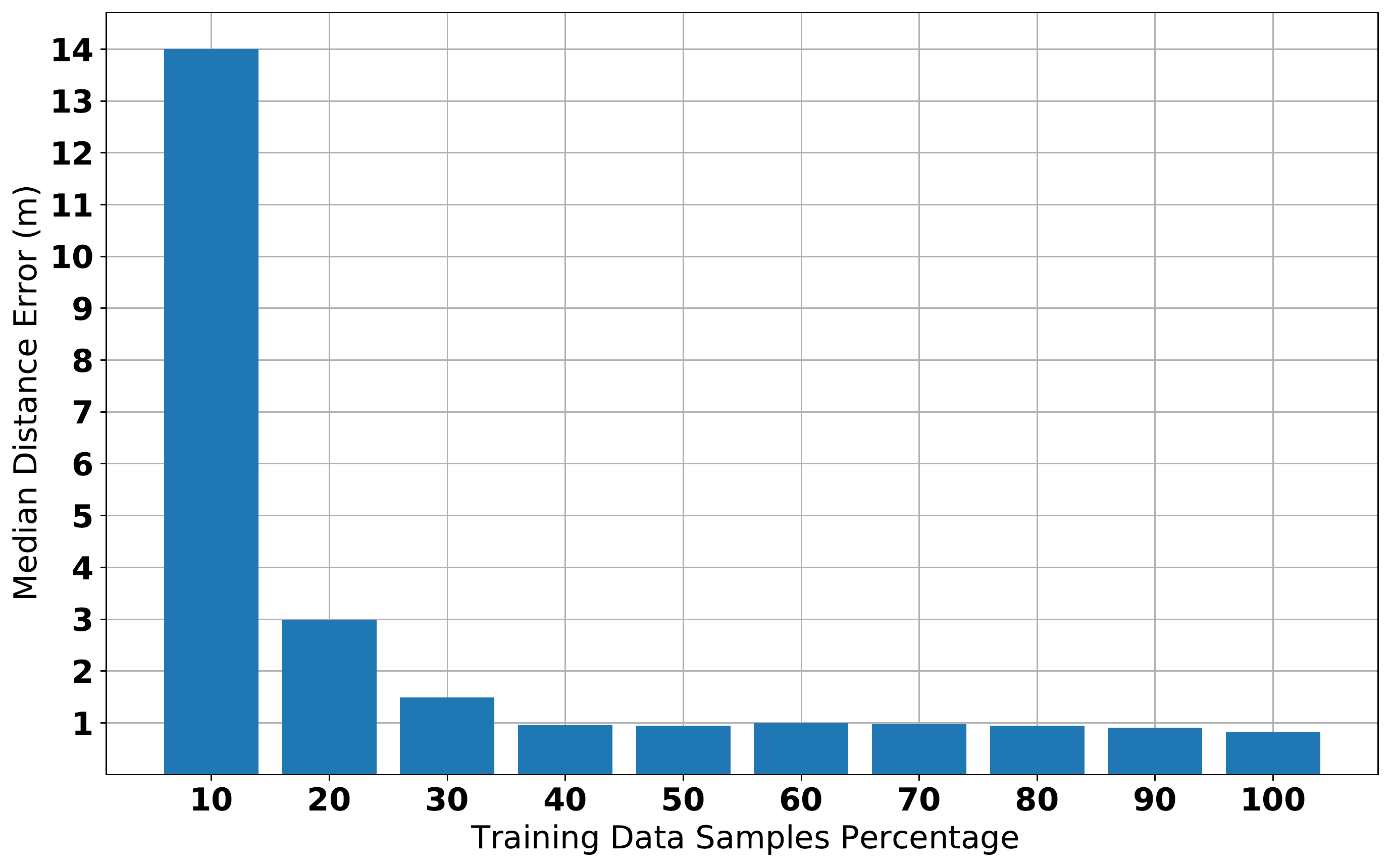}
        \caption{ Effect of reducing the training set by some percentage on system accuracy.}
        \label{fig:Training_Red}   
    \end{minipage}
   \begin{minipage}{0.32\linewidth}
        \centering
        \includegraphics[width=\linewidth,height=5.0cm,]{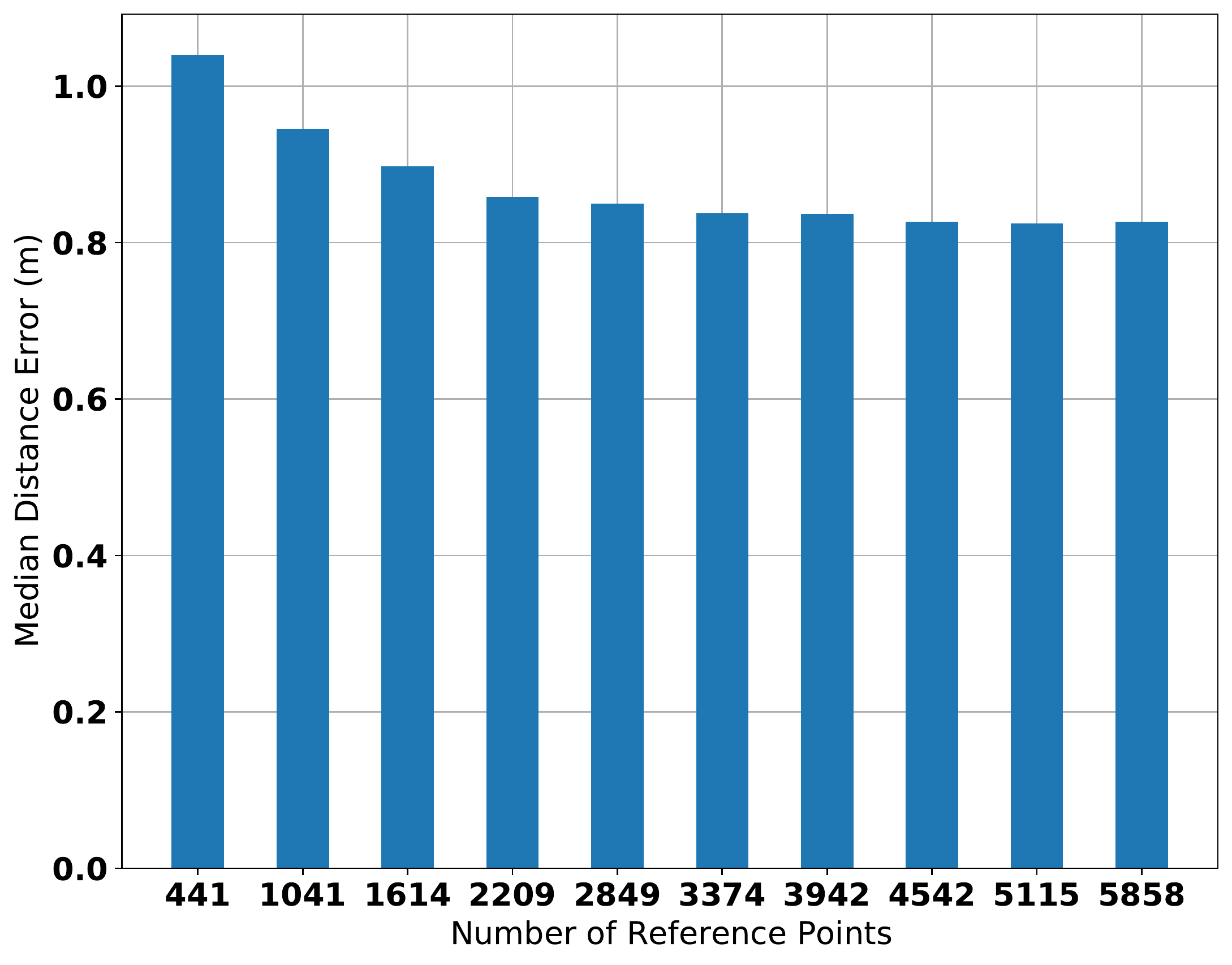}
        \caption{ Effect of reducing the number of reference points on system accuracy.}
        \label{fig:Refrence_points_PER} 
    \end{minipage}
    \begin{minipage}{0.32\linewidth}
        \centering
        \includegraphics[width=\linewidth,height=5.0cm,]{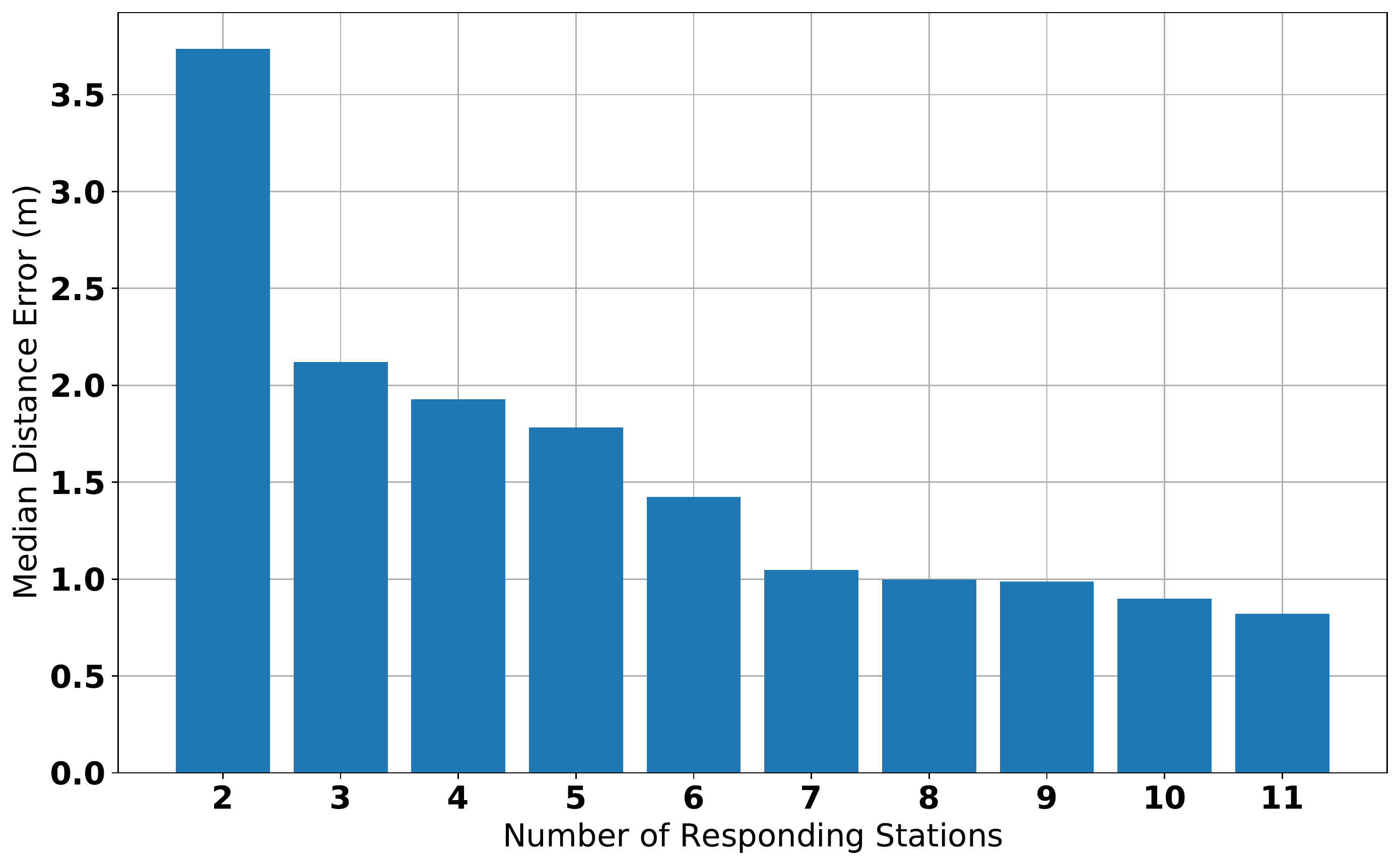}
        \caption{ Effect of Increasing the number of Responding Stations on system accuracy.}
        \label{fig:n_responding}  
    \end{minipage}
\end{figure*}
\begin{figure*}[t]
    \centering
   \begin{minipage}{0.32\linewidth}
        \centering
        \includegraphics[width=\linewidth,height=5.0cm,]{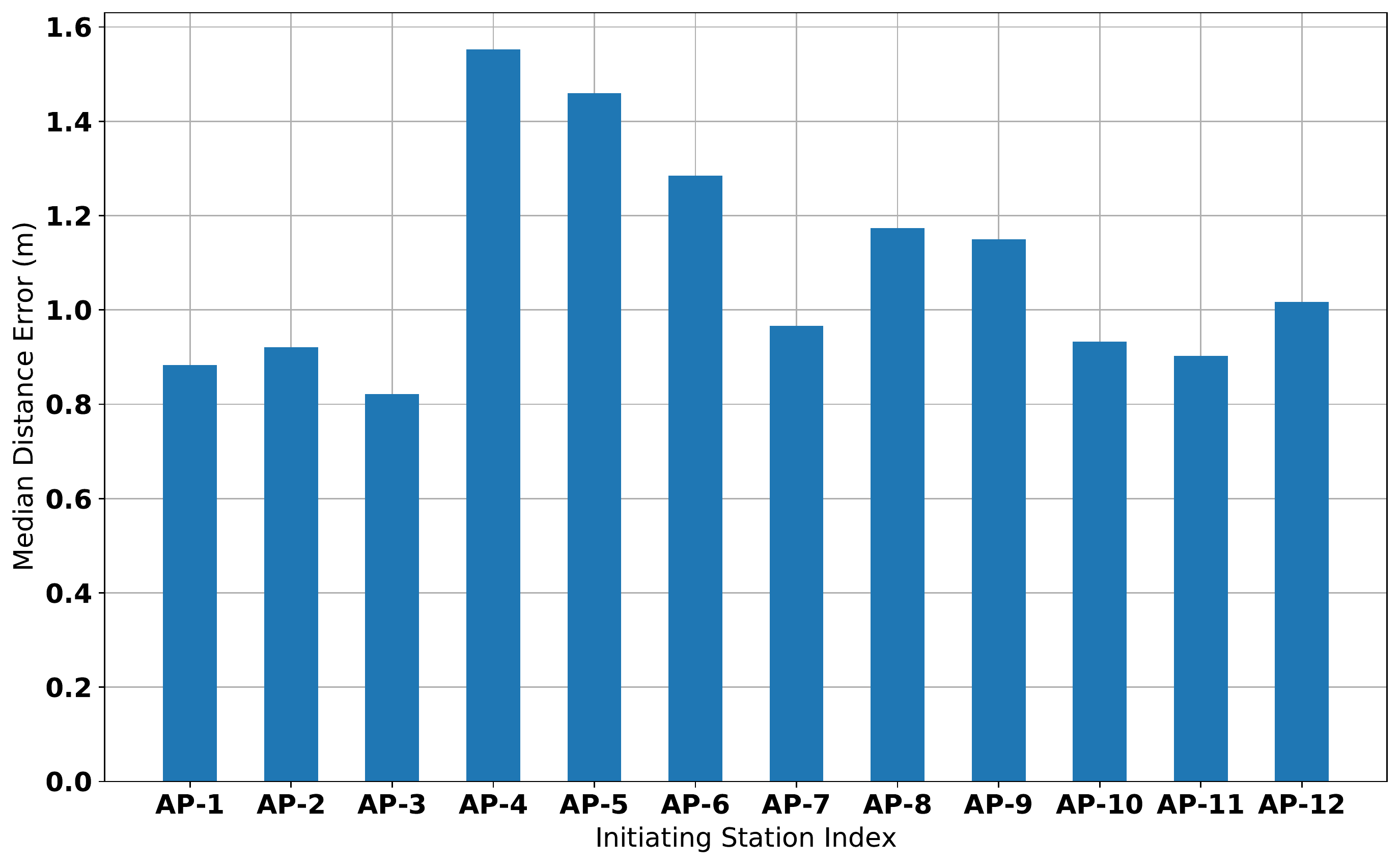}
        \caption{ Effect of Choosing the place of initiating station relative to the responding stations.}
        \label{fig:initiating_indx}  
    \end{minipage}
   \begin{minipage}{0.32\linewidth}
        \centering
        \includegraphics[width=\linewidth,height=5.0cm,]{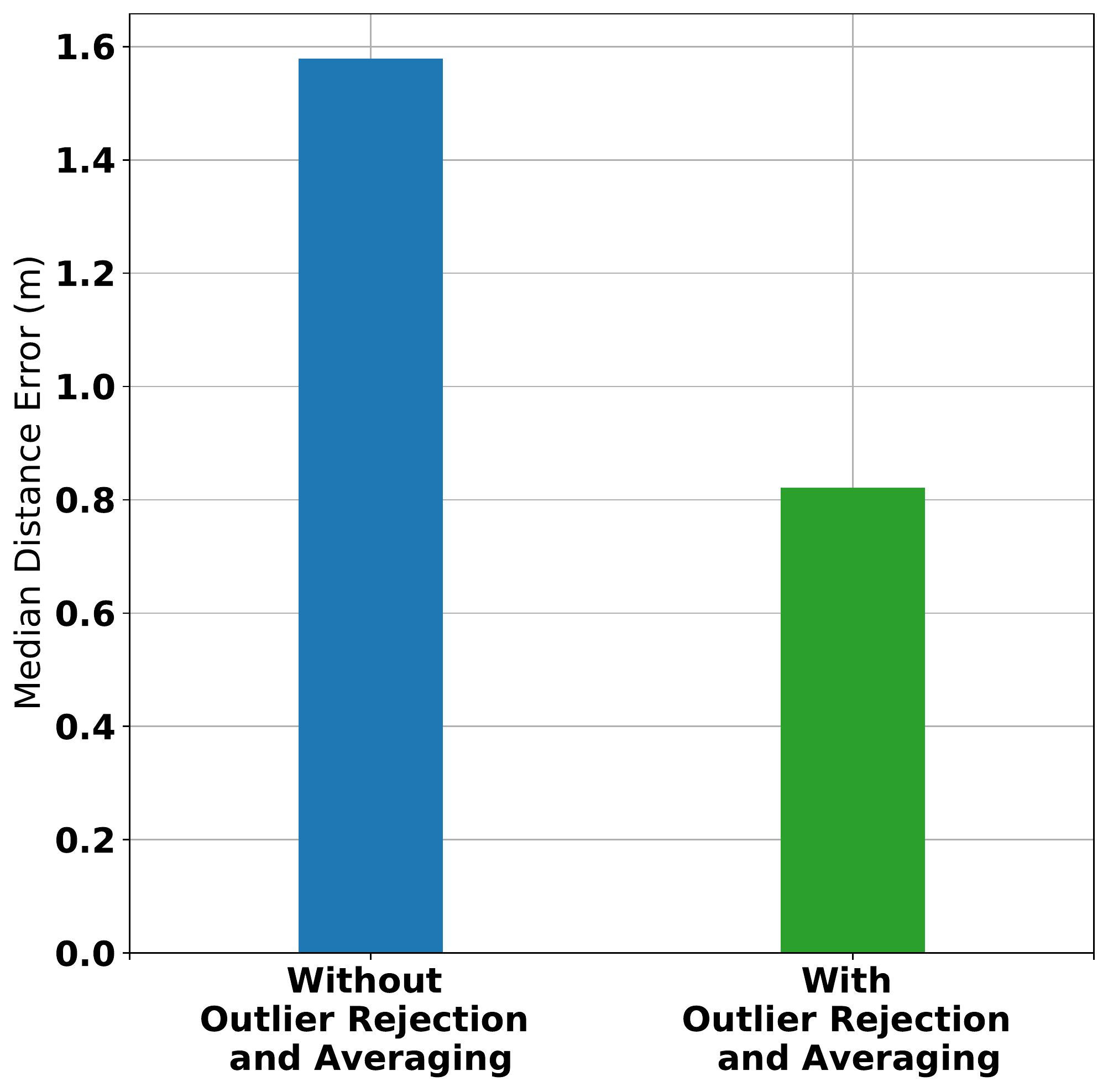}
        \caption{ Effect of post processing stages on system accuracy.}
        \label{fig:Post_Processing}  
    \end{minipage}
     \begin{minipage}{0.32\linewidth}
        \centering
        \includegraphics[width=\linewidth,height=5.0cm,]{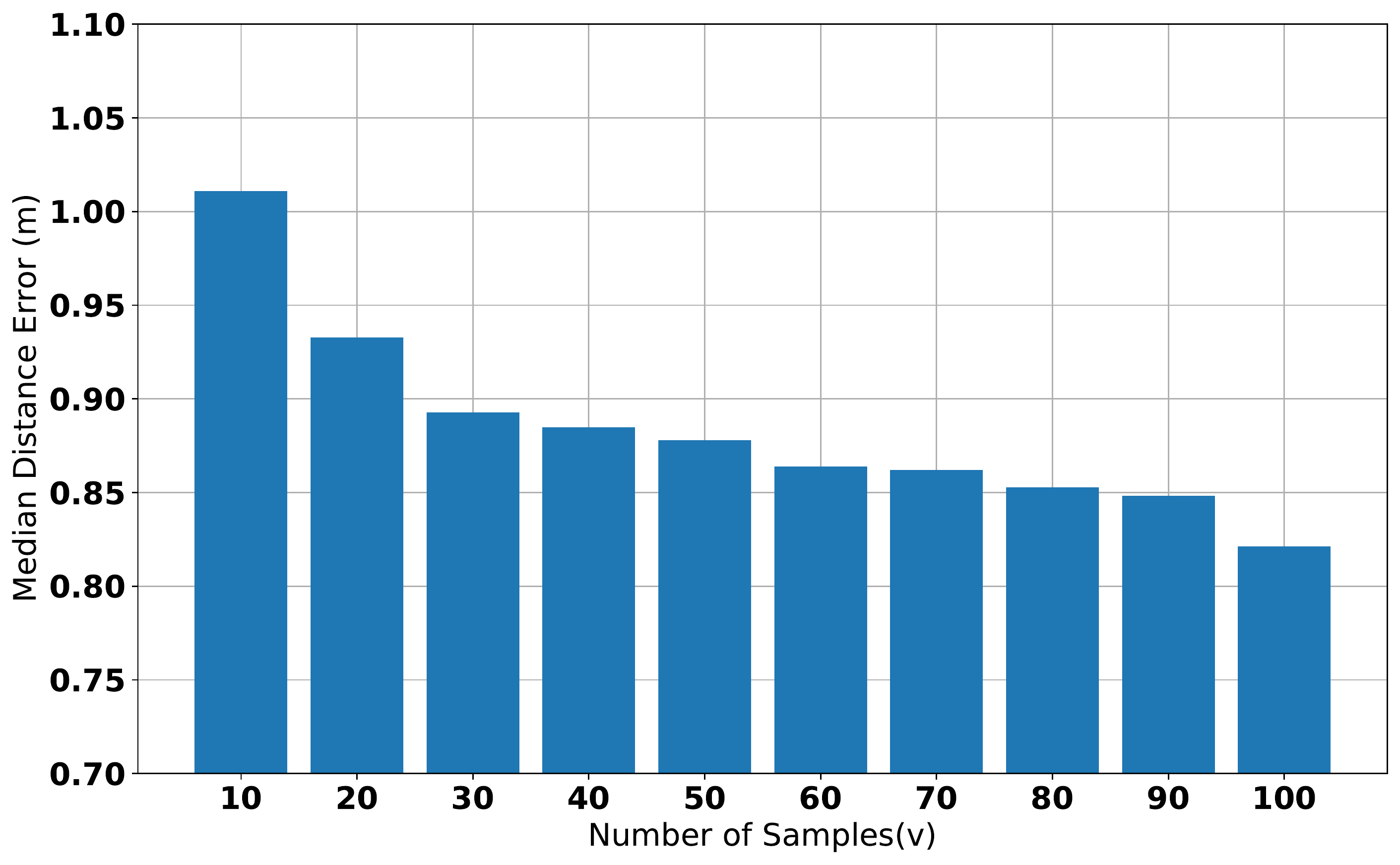}
        \caption{ Effect of varying no. of samples (v) on system accuracy.}
        \label{fig:v_number}
    \end{minipage}
\end{figure*}

\subsubsection{The Size of The Training Set}
Figure~\ref{fig:Training_Red} shows the effect of increasing the number of training samples on the system's accuracy. The figure shows that the more samples in training, the better the localization accuracy. This can be justified due to the diversity of examples that leads to better learning. However, \sys{} maintains a localization accuracy of around 1 meter even with 40\% of the training data. This is due to the learning ability of the model gained due to the regularization techniques. 

   



\subsubsection{Density of Reference Points} In this section, we explore the impact of reference point density on \sys{}'s performance. To do that, we gradually reduced the density of reference points.
The results shown in Figure~\ref{fig:Refrence_points_PER}, indicate the presence of a direct correlation between increased density and improved localization accuracy. The highest accuracy of \sysAcc{}m is achieved when the number of reference points reached 4542, beyond which further density increases do not significantly improve accuracy. This finding supports the idea that a higher density of reference points provides the deep learning model with more data, thereby facilitating better learning of the complex mapping function between the TDoA features and corresponding locations. 

\subsubsection{Number of Responding Stations}
Figure~\ref{fig:n_responding} shows how the accuracy of the system is affected by the number of responding stations in the area. This is implemented by randomly dropping APs from feature vectors. The figure shows that the accuracy gets better as the number of responding stations increases. The more heard APs, the more TDoA values, better discrimination between the different locations, and therefore better localization accuracy.


\subsubsection{Initiating Station Place Relative to Responding Stations}
In this section, we evaluate the impact of the location of the initiating station on the performance of \sys{}. The results presented in Fig.\ref{fig:initiating_indx} reveal a clear correlation between the location of the initiating station and the localization accuracy of \sys{}. The best performance, characterized by the lowest error, was achieved when AP-3 was selected as the initiating station, due to its close proximity to the responding stations, as illustrated in the floor plan of the environment (Figure \ref{fig:Office_2}).
Similarly, access points located in the center of the environment, such as AP-1, AP-2, AP-7, AP-10, and AP-11, provided sub-meter accuracy. On the other hand, selecting the initiating station at the boundary of the environment (e.g., AP-4 or AP-5) led to a drop in accuracy. This is because these access points are farther from the responding stations and, as a result, either become non-detectable or are detected with delay.
These results suggest that the localization accuracy of \sys{} can be optimized by choosing the initiating station closer to the center of mass of all the responding APs' locations. 

\subsubsection{The Online Smoothing Module}
Figure~\ref{fig:Post_Processing} shows the localization accuracy of \sys{} with and without using the online Smoothing Module. The figure depicts the efficacy of the online smoothing of a sequence of consecutive estimates in improving the system accuracy by $92\%$ as well as increasing its robustness to outliers that might occur due to abrupt changes in the environment  (e.g., temporal blockage of the signal LoS caused by a passing person). Additionally, we experimented in Fig.~\ref{fig:v_number} the effect of varying the sequence length on the system accuracy. As shown in the figure, increasing the number of $v$ samples improves the system's accuracy as the availability of several candidates of the user location facilitates outlier detection and removal.

\begin{figure}[t]
 \centering
 \begin{minipage}[b]{0.81\linewidth}
    \includegraphics[width=\linewidth,height=4.7cm,]{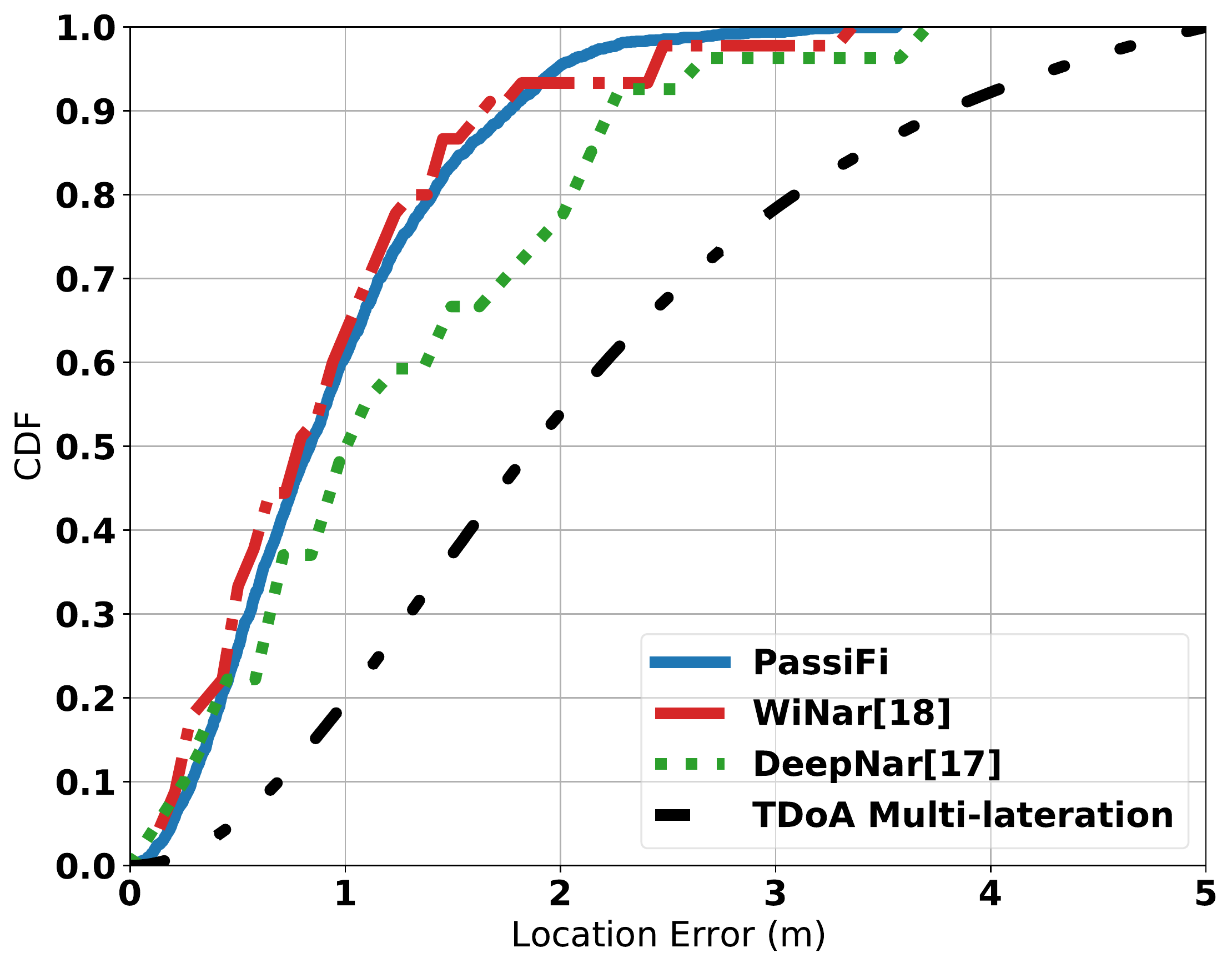}
    \caption{ Comparison of CDFs.}
    \label{fig:CDF}  
 \end{minipage}
 \vspace{-0.6cm}
\end{figure}

\subsection{Comparative Evaluation}
In this section, we compare the performance of \sys{} against TDoA multilateration, WiNar\cite{hashem2020winar}, and DeepNar\cite{hashem2020deepnar} techniques. WiNar\cite{hashem2020winar} relies on RTT data collected at pre-defined points to estimate the user's location through a probabilistic method, while DeepNar\cite{hashem2020deepnar} trains a deep learning-based classification model on RTT fingerprints. The model outputs the likelihood of the user being at one of the predefined reference points.   

Fig.~\ref{fig:CDF} represents the CDF of the localization errors of different techniques on the same testbed. The figure shows that \sys{} achieves a remarkable improvement in median localization accuracy compared to the multilateration technique by $128\%$. 
This can be justified as multilateration mainly relies on the direct path between the receiver and transmitter, which cannot be guaranteed in any indoor environment due to possible blocking of the LoS. However, \sys{} overcomes this challenge by fingerprinting TDoA at the reference locations.
Additionally, \sys{}  also achieves applaudable performance relative to the state-of-the-art active RTT systems with a median sub-meter accuracy of \sysAcc{}m while WiNar\cite{hashem2020winar} and DeepNar\cite{hashem2020deepnar} achieve 0.79m and 0.9m, respectively.
However, what truly distinguishes \sys{} is not solely its localization accuracy but also its passive operation, which prioritizes the user's privacy and shields them from potential attacks.


\section{Conclusion} \label{Conclusion}
In this paper, we presented \sys{}, a deep learning-based indoor localization system that integrates fingerprinting, and time-based techniques in a privacy-aware manner. The system passively collects Wi-Fi signals and leverages the TDoA of the received signals as  features to train a deep neural network model that provides sub-meter level localization accuracy.
The evaluation results demonstrate the superior performance of \sys{} compared to existing techniques, with a remarkable improvement of at least $128\%$ in accuracy. Additionally, the privacy-preserving design of \sys{} protects users from the risk of being localized by malicious actors and safeguards the integrity of their measurement data. In future work, we aim to assess and boost the system's scalability.

\section*{Acknowledgment}
This work was partially funded by the Japan Society for the Promotion of Science, KAKENHI Grant number 22K12011, and NVIDIA award.

\IEEEpeerreviewmaketitle

\ifCLASSOPTIONcaptionsoff
  \newpage
\fi

\bibliography{__Ref}
\bibliographystyle{ieeebib}

\end{document}